\documentclass[journal]{IEEEtran}
\usepackage{graphicx}
\usepackage{amsmath}
\usepackage{amssymb}
\usepackage{algpseudocode}
\usepackage{multirow}
\usepackage{cite}
\usepackage{enumitem}
\usepackage[colorlinks=true, citecolor=green, linkcolor=red]{hyperref}
\usepackage[linesnumbered,ruled]{algorithm2e}

\newcommand{\ie}{\emph{i.e.}}
\newcommand{\eg}{\emph{e.g.}}

\ifCLASSINFOpdf
\else

\fi

\begin{document}
%

\title{Biphasic Face Photo-Sketch Synthesis via Semantic-Driven Generative Adversarial Network with Graph Representation Learning}
%
%
%
\author{Xingqun Qi$^{\ast}$, Muyi Sun$^{\ast}$, Zijian Wang, Jiaming Liu, Qi Li,~\IEEEmembership{Member,~IEEE}, Fang Zhao, \\ Shanghang Zhang, Caifeng Shan$^{\dagger}$,~\IEEEmembership{Senior Member,~IEEE}

\thanks{This work was supported by the Talent Introduction Program for Youth Innovation Teams of Shandong Province and also in part supported by the National Natural Science Foundation of China No.62306309.}
\thanks{Xingqun Qi is with the Academy of Interdisciplinary Studies, Hong Kong University of Science and Technology. (e-mail: xingqun.qi@connect.ust.hk).}
\thanks{Muyi Sun and Qi Li are with the CRIPAC, NLPR, Institute of Automation, Chinese Academy of Sciences, Beijing 100190, China. Muyi Sun is also with the School of Artificial Intelligence, Beijing University of Posts and Telecommunications, Beijing 100876, China (e-mail: muyi.sun@cripac.ia.ac.cn; qli@nlpr.ia.ac.cn)
)}
\thanks{Jiaming Liu and Shanghang Zhang are with the National Key Laboratory for Multimedia Information Processing, School of Computer Science, Peking University.
(e-mail: jiamingliu@stu.pku.edu.cn, shanghang@pku.edu.cn).}
\thanks{Fang Zhao is with the School of Intelligence Science and Technology, Nanjing University, Nanjing 210023, China (e-mail: zhaofang0627@gmail.com).}
\thanks{Caifeng Shan is with the College of Electrical Engineering and Automation, Shandong University of Science and Technology, Qingdao 266590, China and also the School of Intelligence Science and Technology, Nanjing University, Nanjing 210023, China. (Email: caifeng.shan@gmail.com)}
\thanks{${\ast}$ These authors contributed equally to this work. This work is done when Xingqun Qi was an intern at the CRIPAC, NLPR, Institute of Automation, Chinese Academy of Sciences, and also with the School of Computer Science, Peking University. }
\thanks{$^{\dagger}$\textbf{Corresponding author: Caifeng Shan.}}

}

\markboth{Journal of \LaTeX\ Class Files,~Vol.~14, No.~8, August~2015}%
{Shell \MakeLowercase{\textit{et al.}}: Bare Demo of IEEEtran.cls for IEEE Journals}

\maketitle

\begin{abstract}
Biphasic face photo-sketch synthesis has significant practical value in wide-ranging fields such as digital entertainment and law enforcement. Previous approaches directly generate the photo-sketch in a global view, they always suffer from the low quality of sketches and complex photo variations, leading to unnatural and low-fidelity results. In this paper, we propose a novel Semantic-Driven Generative Adversarial Network to address the above issues, cooperating with Graph Representation Learning. Considering that human faces have distinct spatial structures, we first inject class-wise semantic layouts into the generator to provide style-based spatial information for synthesized face photos and sketches. Additionally, to enhance the authenticity of details in generated faces, we construct two types of representational graphs via semantic parsing maps upon input faces, dubbed the IntrA-class Semantic Graph (IASG) and the InteR-class Structure Graph (IRSG). Specifically, the IASG effectively models the intra-class semantic correlations of each facial semantic component, thus producing realistic facial details. To preserve the generated faces being more structure-coordinated, the IRSG models inter-class structural relations among every facial component by graph representation learning. To further enhance the perceptual quality of synthesized images, we present a biphasic interactive cycle training strategy by fully taking advantage of the multi-level feature consistency between the photo and sketch. Extensive experiments demonstrate that our method outperforms the state-of-the-art competitors on the CUFS and CUFSF datasets.

\end{abstract}


\IEEEpeerreviewmaketitle

\section{Introduction}

\IEEEPARstart{B}{iphasic} face photo-sketch synthesis refers to generating sketches from face photos and, conversely, generating photos from face sketches.
The wide-ranging application fields of the biphasic face photo-sketch synthesis include digital entertainment, law enforcement, and criminal case judgment. 
Specifically, face sketch is one of the most popular and fundamental portrait painting styles in the scope of digital entertainment~\cite{b1}. 
In law enforcement and criminal case judgment, police commonly just hold the sketches of suspects drawn from the description of the witnesses. 
Face photos synthesized from these sketches with clear identities and manifest features can provide a feasible way to promote the efficiency of justice criminal cases~\cite{b2}. 
However, it requires a vast time and effort to create distinct face sketches by professional artists. 
Due to the vital practical value, it is especially essential to automatically synthesize face photos and sketches with realistic effects and consistent identity preservation. 

\begin{figure}
\begin{center}
\includegraphics[width=0.95\linewidth]{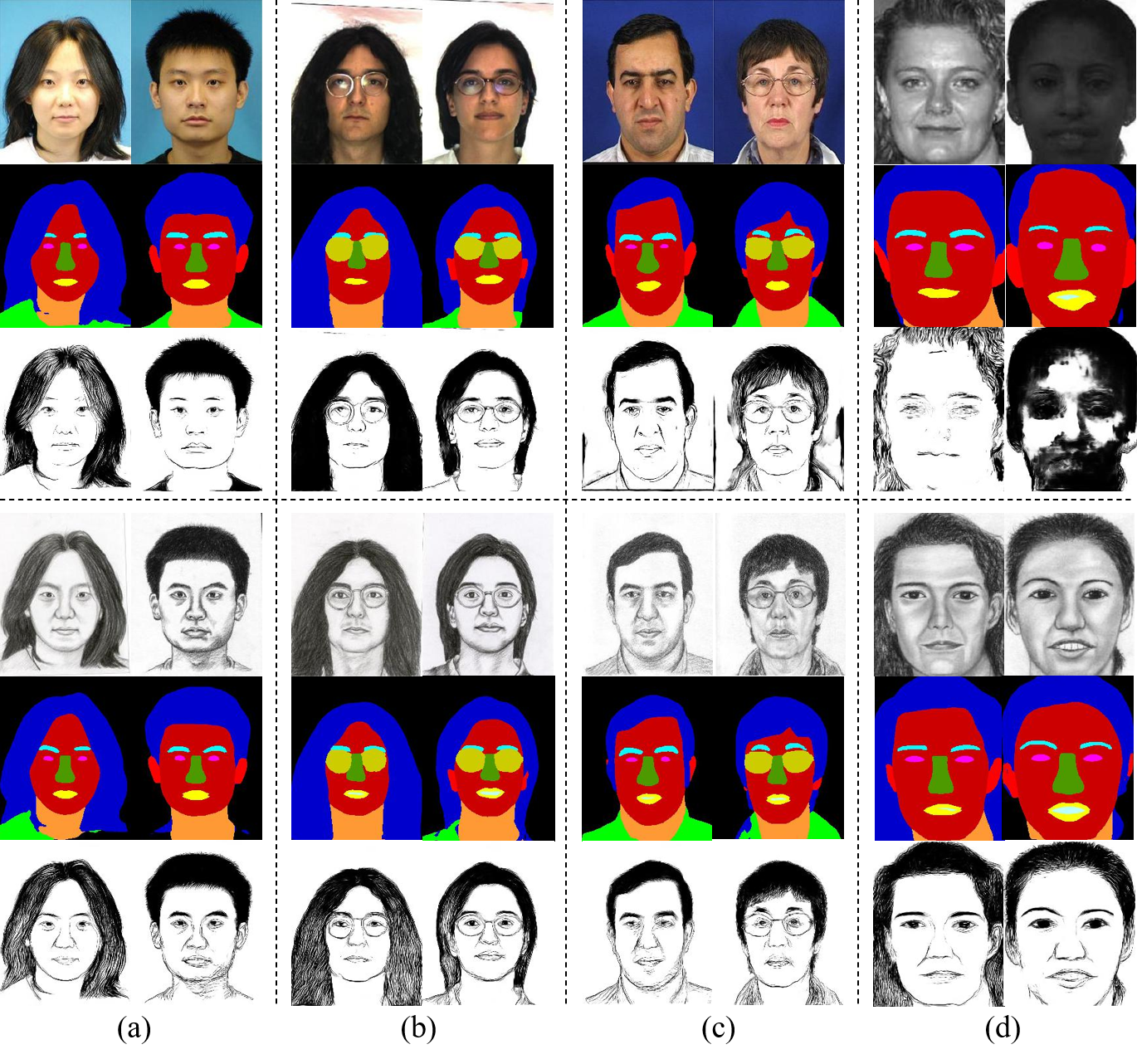}
\end{center}
   \caption{Illustrations of samples with facial prior knowledge and parsing layout in different databases:  (a) CUHK database, (b) AR database, (c) XM2VTS database, (d) CUFSF database. For each database, from top to bottom are face photos, the parsing layout of photos, the saliency detection result of photos, sketches, the parsing layout of sketches, and the saliency detection result of sketches, sequentially.
}
\label{fig:prior}
\end{figure}



Numerous approaches have been proposed to address this task, which can be roughly grouped into exemplar-based approaches~\cite{b3}, linear-regression based approaches~\cite{b4}, and generative-model based ones~\cite{b5}.
Exemplar-based approaches~\cite{b1,b3,b6,b7,b8} focus on bridging the mapping from photo to sketch via limited exemplar-paired patches, leading to over-smoothed and low personal identity results.
Besides, linear-regression based approaches~\cite{b9,b10,b11,b12} always directly construct the linear mapping between photos and sketches. 
Recently developed generative-based methods~\cite{b14,b15,b16,b17,b18,b19,b20} show better performance compared with the other two types. 
However, although these methods could generate sketches or face photos automatically, they adopted a holistic manner for generation and overlook the local specificity of different facial regions. Therefore, the face directly generated by these methods might yield distortion and noises in detailed parts. Meanwhile, the fidelity of the image and the consistency of the face identity also need to be improved.

Motivated by the above studies, we propose a novel Semantic-Driven Generative Adversarial Network with Graph Representation Learning for biphasic face photo-sketch synthesis. 
Our model is built upon the key insight that the human face has obvious spatial structures that determine the fidelity and identity consistency of face images.
Therefore, the structural information should be enhanced by incorporating external prior or learned by designing specific objectives during the training process. 
In this paper, we design three methods to incorporate and learn structural information: saliency detection based guidance, face semantic injection, and two graph-based training objectives.

The saliency detection guidance utilizes the face saliency map as prior input to our network. Thanks to the recently advanced face saliency detection technique~\cite{b21}, we first leverage a pre-trained saliency detector to obtain the saliency maps, which incorporate overall facial structural information by identifying the most conspicuous or prominent parts of face images.
As displayed in Fig~\ref{fig:prior}, it is worth noting that the saliency map is different from the corresponding target sketch since the sketch contains more personal styles and textures of the painter \eg, the shadows and outlines.

Then, considering the generated results should be structure consistent with the input human face, we perform face semantic injection in the network. 
Specifically, we transform the class-wise semantic layouts into two modulation parameters and inject them into the network decoder to provide style-based spatial information.
In this fashion, the facial structure is well-preserved as well as the person's identity is effectively guaranteed during the generation\footnote{This is essential in sketch-to-photo synthesis when facing the criminal case judgment.}. 
Here, we utilize a superior face parser to extract accurate semantic layouts for both photo and sketch, as depicted in Fig.~\ref{fig:prior}.


To ensure the details of generated results are authentic, we propose two training objectives, which employ two 
novel graph representations for measuring facial structured information, dubbed the \textbf{I}ntr\textbf{A}-class \textbf{S}emantic \textbf{G}raph (IASG) and the \textbf{I}nte\textbf{R}-class \textbf{S}tructure \textbf{G}raph (IRSG). In particular, IASG leverages the human face semantic layouts as guidance to model the intra-class correlation of each facial component represented by a graph node. Here, we calculate the graph node as the mean center and variance center of the corresponding facial components. Considering that different facial components always contain large-range pixel numbers, we present an adaptive re-weighting algorithm to balance the contribution of detailed facial parts. Then, intra-class correlation can be effectively expressed by the similarity between the mean center (or variance center) and each pixel in the corresponding facial component. By enforcing the generated fake IASG to be consistent with the target ones, the details in the synthesized images achieve high-fidelity effects. 

As a complement, the IRSG aims to keep the generated face photo and sketch more structure-coordinated with targets, globally. IRSG models the inter-class relations among every facial component. Concretely, the inter-class relations are built by computing the affinity scores between every two different mean centers (and variance centers). Similar to IASG, we utilize the target IRSG to supervise the synthesized fake ones, thus producing the natural human faces.
 
Moreover, based on the observation that the paired photo-sketch shared many personal characters, we design a novel biphasic iterative cycle training strategy to improve the perceptual quality of synthesized images. Here, we first train two models of photo generation and sketch generation, respectively. Then, we exploit the pre-trained sketch generation model as a personal knowledge extractor to boost photo generation. Iteratively, the sketch generation model is improved with the help of a pre-trained photo personal knowledge extractor. In this fashion, our framework enables high-quality biphasic face photo-sketch synthesis through multi-stage iterations. Extensive experiments demonstrate that our method significantly outperforms various counterparts, displaying natural and realistic facial details.




The main contributions of this study are summarized as:
\begin{itemize}[leftmargin=*]
\item We propose a novel Semantic-Driven Generative Adversarial Network with Graph Representation Learning for generating realistic photos and distinct sketches.
\item We construct two types of representational graphs and design corresponding constraints to facilitate the preservation of the details and coordinate structures in generated face photos and sketches.
\item We propose a novel biphasic iterative cycle training to improve the perceptual quality of the synthesized images by effectively taking advantage of the multi-level feature consistency between the photo and sketch.
\item Extensive comparison experiments are conducted on CUFS and CUFSF datasets, showing our method obtains state-of-the-art performance.
\end{itemize}

Compared to our preliminary work in \cite{b22}, the improvements and extensions are consulted in three-fold: 1) We propose a novel inter-class structure graph that facilitates the preservation of the details and personal identities in generated face photos and sketches; 2) A novel biphasic iterative cycle training strategy is proposed to improve the perceptual quality of synthesized images; 3) We conduct the extension experiments on both sketch synthesis and face photo synthesis tasks. Driven by the aforementioned improvements, our framework achieves superior performance on \textbf{biphasic} photo-sketch synthesis in a \textbf{unified} manner. This significantly facilitates the application community on digital entertainment and law enforcement.

\section{Related Work}

\subsection{Biphasic Face Photo-Sketch Synthesis}
Biphasic face photo-sketch synthesis has developed rapidly in the last few decades. Massive works have been proposed to solve this problem, which includes two closely related subtasks: face sketch synthesis and face photo synthesis. Therefore, researchers often analyze and discuss these two subtasks in a unified manner. 



Deep neural network based approaches are the mainstream routine of biphasic photo-sketch synthesis in recent years, which have gradually emerged with the boost of Generative Adversarial Networks. Ji \emph{et al.} \cite{b16} employed multi-domain adversarial methods to construct a mapping from photo-domain to sketch-domain. Zhu \emph{et al.} \cite{b19} borrowed knowledge from transfer learning and proposed a lightweight network supervised by a high-performance larger network. Zhang \emph{et al.} \cite{b23} embedded the photo parsing priors and designed a parametric sigmoid activation function in GAN based framework to facilitate robust sketch synthesis. Recently, Yu \emph{et al.} \cite{b20} decomposed the face parsing layouts into multiple compositions and encoded them into conditional GAN (cGAN) for biphasic face photo-sketch synthesis which achieved state-of-the-art performance. Moreover, Duan \emph{et al.} \cite{b18} introduced the gradient-based self-attention mechanism to combine the global residual connection and local residual connection in the proposed network which achieved better results. Besides, Zhang \emph{et al.} \cite{b24} designed a dual transfer strategy to promote the biphasic face photo-sketch synthesis. Lin \emph{et al.} \cite{b25} proposed the feature injection module to preserve the identity of synthesized sketches and photos. However, the sketches and photos synthesized by these methods are short of realistic detailed depictions.

Inspired by previous works, we inject the Semantic information into the generator of our proposed network. However, our injection module is different from previous works which embedded the multi-level identity feature into the network. In contrast, we aim at providing class-wise style-based spatial supervision for synthesized face photos and sketches. Furthermore, the previous work \cite{b25} did not follow the common experiment setting which ignored the influence of background on the generated results.

\begin{figure*}[t]
\begin{center}
\includegraphics[width=0.9\linewidth]{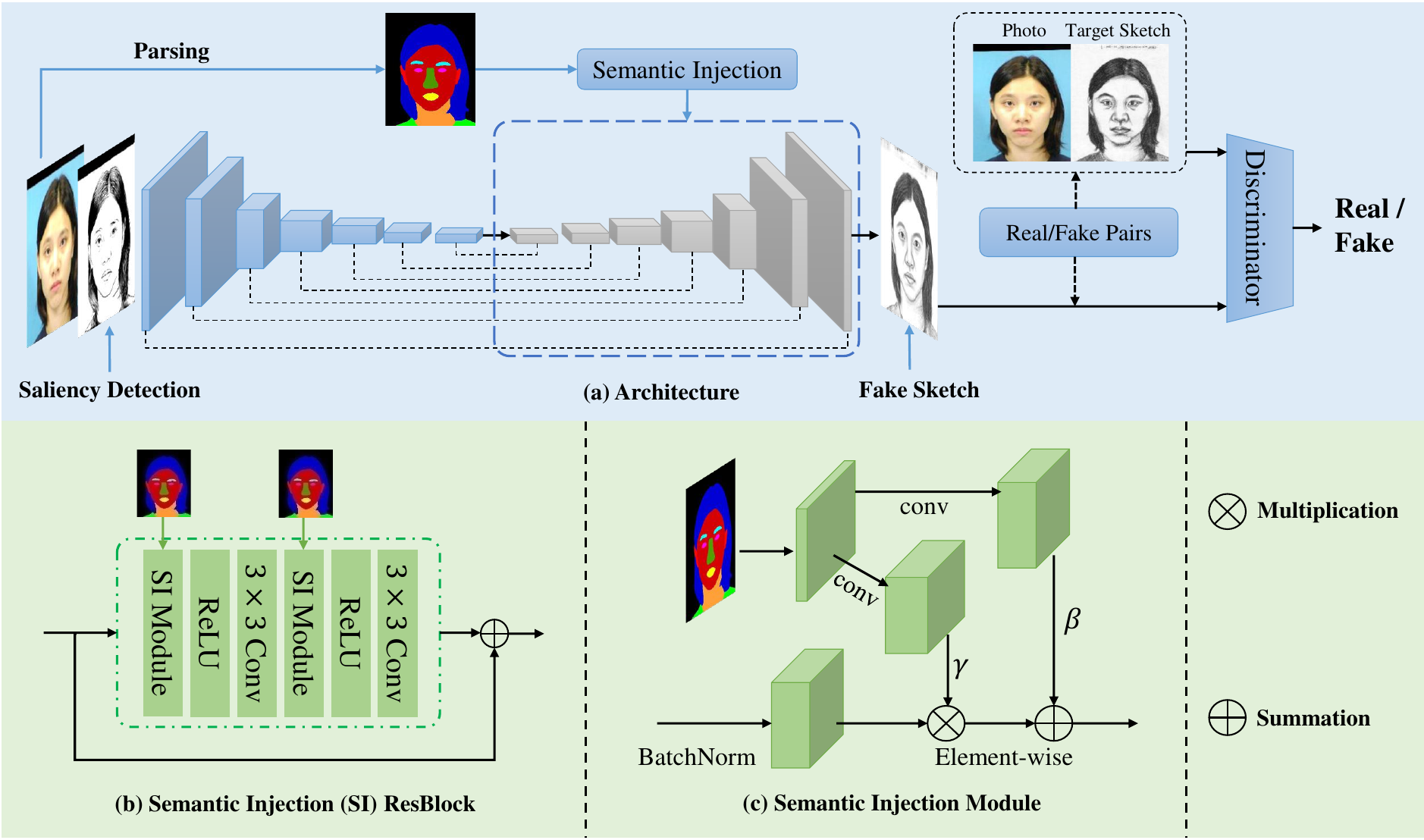}
\end{center}
   \caption{The pipeline of our semantic-driven network. (a) The over architecture of the network exploits Pix2Pix as the backbone. The decoder contains several Semantic Injection (SI) ResBlocks with upsampling layers. (b) Details of Semantic Injection (SI) ResBlocks. (c) Details of Semantic Injection Module. In each Semantic Injection Module, the semantic map is convoluted to produce pixel-level normalization parameters $\gamma $ and $\beta $.}
\label{fig:architecture}
\end{figure*}

\subsection{Paired Image-to-Image Style Transfer}
Paired image-to-image style transfer tasks can be regarded as a combination of image-to-image translation tasks and style transfer tasks. The image-to-image translation is often formulated as pixel-wise image generation tasks applied with paired images like biphasic face photo-sketch synthesis. Isala \emph{et al.} \cite{b26} proposed a cGAN architecture to solve the image-to-image translation task with paired input and output named Pix2Pix. Due to the eminent performance of the Pix2Pix on the paired dataset, researchers have made numerous improvements based on Pix2Pix and applied them to a wide range of research fields \cite{b27,b28, b67, b68}. By combining Pix2Pix and residual blocks, Wang \emph{et al.} \cite{b29} proposed a novel network architecture to generate high-resolution images named pix2pixHD. Moreover, Park \emph{et al.} \cite{b30} introduced the semantic layouts as spatial supervision injected in the pix2pixHD for synthesizing photorealistic images. Motivated by previous research, we exploit cGAN like Pix2Pix as our backbone network.

The biphasic face photo-sketch synthesis task can be treated as an image style transfer task between realistic photos and vivid sketch portraits. Gatys \emph{et al.} \cite{b31} successfully applied pre-trained CNNs to the image style transfer task. Furthermore, Ulyanov \emph{et al.} \cite{b32,b33} optimized the style transfer process by manipulating the Batch Normalization (BN) layers and Instance Normalization (IN) layers. Dumoulin \emph{et al.} \cite{b34} utilized a group of parameters to realize the transfer of various image styles. Consecutively, Huang \emph{et al.} \cite{b35} proposed the adaptive instance normalization (AdaIN) layers which could perform arbitrary style transfer without training repeatedly. Recently, Park \emph{et al.} \cite{b13} put forward the spatially-adaptive normalization (SPADE) layers that inject the image style from the semantic layouts to obtain photorealistic images. Motivated by the previous research, we inject the style-based statistic information into the network to generate images with more distinct characteristics.

\subsection{Graph Representation Learning}
Graph representation learning plays a significant role in computer vision tasks which could encode each node represented by a low-dimensional dense embedding in the graph structure \cite{b36}. Perozzi \emph{et al.} \cite{b37} effectively embedded information of nodes by randomly sampling the graph structure. Furthermore, Defferrard \emph{at al.}\cite{b38} designed a novel pooling strategy to rearrange the nodes for preserving more useful information of graph. Li \emph{et al.} \cite{b39} constructed the structural graph of actions to capture the high-order dependencies between successive skeletons in the proposed actional-structural graph convolution network for skeleton-based action recognition. Besides, Ren \emph{et al.} \cite{b40} adopted the graph generator to build the connections among spatial parts and construct the feature graph of these nodes representation for biometrics. Yang \emph{et al.} \cite{b41} utilized a novel graph representation specifically designed for sketches by bridging the structural hierarchical relationship of sketches. This work has significantly improved the recognition accuracy of sketches.
Wu \emph{et al.} \cite{b65} propose an adaptive graph representation learning scheme for video person Re-ID, which enables the contextual interactions between relevant regional features. Jin \emph{et al.} \cite{b66} introduce a self-supervised approach to learning graph node representations by enhancing Siamese self-distillation with multi-scale graph representation learning. As for the biphasic face photo-sketch synthesis task, Zhu \emph{et al.} \cite{b7} combined graphical exemplar-based features with deep neural networks to synthesize high-quality sketches. The results synthesized by \cite{b7} are robust against lighting variations and clutter backgrounds. However, they still suffer from over-smoothing and lacking specific-identify characters in the synthesized photos and sketches. On the contrary, our graph representation learning algorithms can make restraint of intra-class features and inter-class features to help keep the realistic personal details of synthesized photos and sketches.



\section{Method}

In this section, details about our proposed Semantic-Driven Generative Adversarial Network with Graph Representation Learning are presented. First, we introduce the preliminaries and problem formulation of our method. Afterward, the network architecture is described. Subsequently, we elaborate on the graph representation learning algorithms and biphasic iterative cycle training strategy. Finally, the objective functions of our model are introduced.

\subsection{Preliminaries}
Our semantic-driven network aims to construct a biphasic mapping of paired photo-sketch by utilizing class-wise semantic layouts as guidance. Previous researchers directly synthesize sketches or photos in a holistic manner, leading to unnatural and low-fidelity details in results. Our key insight is based on the observation that the human face has a distinct spatial structure. Here, we leverage the class-wise semantic layouts to model this structure information via graph representation learning. Specifically, the \textbf{I}ntr\textbf{A}-class \textbf{S}emantic \textbf{G}raph (IASG) and the \textbf{I}nte\textbf{R}-class \textbf{S}tructure \textbf{G}raph (IRSG) are customized designed. Moreover, we present a novel biphasic iterative cycle training strategy to enhance the perceptual quality of the synthesized images. Besides, we conduct facial saliency detection on the input images to provide overall prior information on facial structure. For brevity, we take the sketch synthesis task as the prototype to introduce the details.


Given paired photo-sketch training samples $\begin{Bmatrix}
\left ( x_{i},y_{i} \right )|x_{i}\in X, y_{i}\in Y
\end{Bmatrix}_{i=1}^{N}$, where $x_{i}$ represents photo and $y_{i}$ represents sketch. The purpose of face sketch synthesis is to construct a mapping from the source photo domain $X$ to the target sketch domain $Y$. As illustrated in Fig. \ref{fig:prior}, we find that there are complex variations in the source domain, resulting in severe impacts on the identity and fidelity of the generated sketches as depicted in Fig. \ref{fig:CUFS} (c), (d) and Fig. \ref{fig:CUFSF} (c), (d). Conversely, in the photo generation task, the low-quality sketches of the source domain would affect the clarity of the generated image. Advanced face saliency detectors effectively capture the global structure information of the input photos or sketches. Regarding this, we first utilize face saliency detection results as prior information to provide the global facial structure. We concatenate the saliency map $M$ and the face photo as the input to the generator. Besides, we also employ the pre-trained face parser to acquire semantic layouts $S$ as guidance. Then, we inject this semantic information into our network to produce the final synthesized result. Therefore, the overall mapping can be formulated as $\begin{Bmatrix}X,M,S\end{Bmatrix}\rightarrow Y$.

\subsection{Semantic Driven Network Architecture}

Fig. \ref{fig:architecture} illustrates the overall architecture of our network. 
First, we concatenate the paired saliency map and face photo as input.
Pix2Pix \cite{b32} is exploited as the backbone including 7 convolutional and downsampling layers in the encoder part. 
Inspired by \cite{b36}, we further design 7 Semantic Injection (SI) ResBlocks in the decoder to strengthen the conditional structure information guided by parsed semantic layouts.
Each ResBlock contains two convolutional layers, two ReLU layers, and two Semantic Injection (SI) modules.

As shown in Fig. \ref{fig:architecture} (c), the SI module takes two inputs: the forward activation features after the Batch Normalization layer and semantic layouts obtained by pre-trained BiSeNet \cite{b42}. In order to prevent semantic ambiguity, we divide the human face into 12 spatial components: two eyes, two eyebrows, two ears, glasses, upper and lower lips, inner mouth, hair, nose, skin, neck, cloth, and background.
Therefore, we have $S= \begin{Bmatrix}s^{(1)},\cdots,s^{(c)} \end{Bmatrix} \in \mathbb{R}^{h\times w\times c}$, where $c\in \begin{bmatrix}1,2,\cdots,12\end{bmatrix}$, $s^{(c)}\in \begin{bmatrix}0,1\end{bmatrix}$, $h$ and $w$ denote the height and width of the semantic maps.
Here, to inject the spatial structure information into the SI module, we perform the convolutional operation on semantic layouts. Thus, two modulation parameters $\gamma $ and $\beta $ are produced to normalize the final output. These two parameters encode sufficient spatial structure information. Then, we conduct multiplication and addition between these two modulation parameters and the normalized activation maps in an element-wise pattern as shown in Fig. \ref{fig:architecture} (c). In this fashion, the facial structure of the synthesized sketch is well-preserved with the input photo. Meanwhile, since the semantic layouts are robust to the complex variations of background, the generated sketches achieve the high-fidelity effect.
Finally, we adopt a patch-wise discriminator to enforce the generated sketches keeping realism.


\begin{figure}[t]
\begin{center}
\includegraphics[width=1\linewidth]{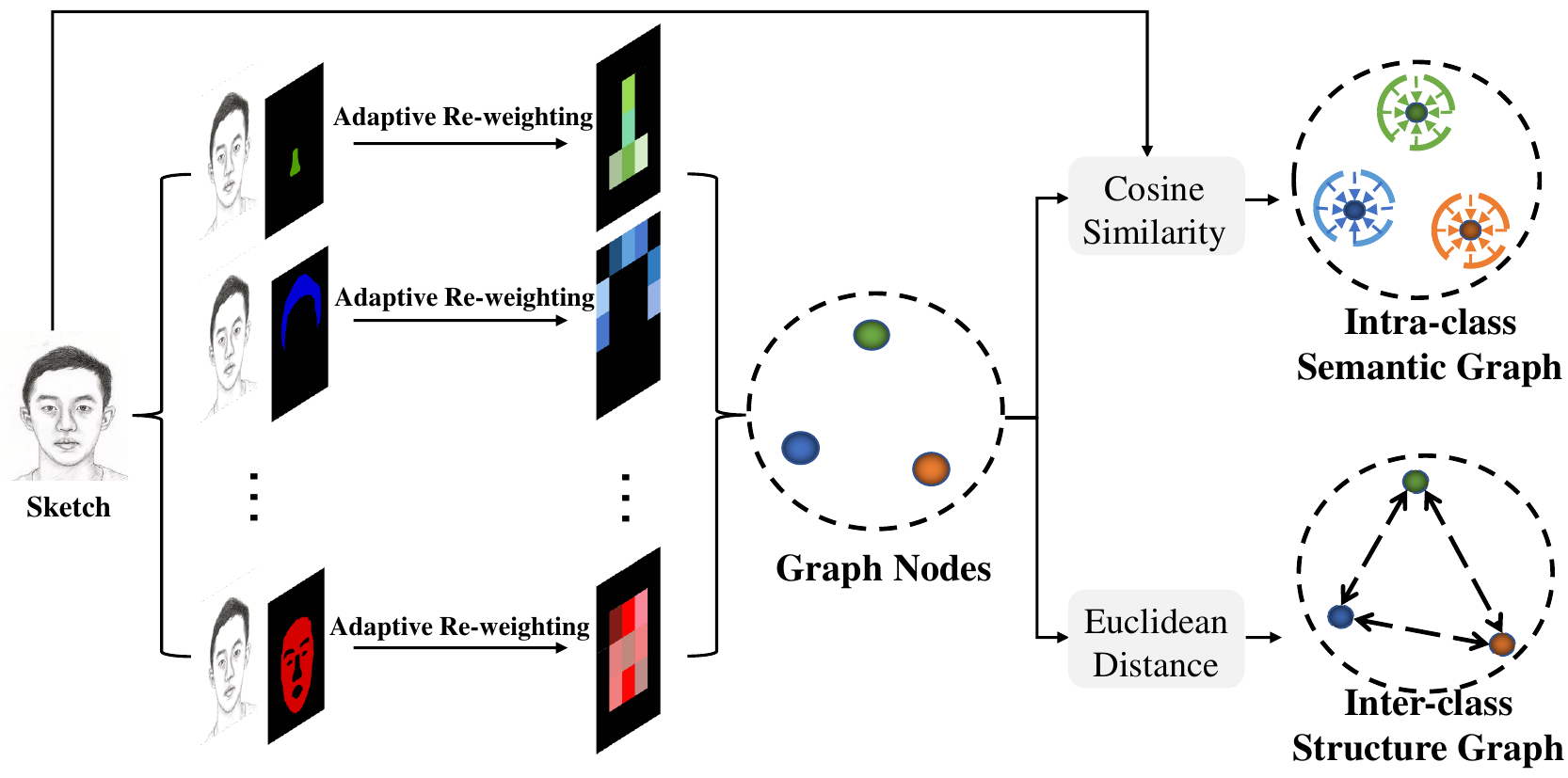}
\end{center}
   \caption{Illustration of our proposed intra-class semantic graph and inter-class structure graph. The semantic masks are produced by a pre-trained face parsing network from the input photo. Besides, there are two types of graph nodes that represent the mean center and variance center of each facial component class.}
\label{fig:graph}
\end{figure}

\subsection{Graph Representation Learning}
Previously, researchers always impose global supervision on the entire generated sketches, resulting in defective performances in facial details. Thus, we construct two representational graphs named the \textbf{I}ntr\textbf{A}-class \textbf{S}emantic \textbf{G}raph (IASG) and the \textbf{I}nte\textbf{R}-class \textbf{S}tructure \textbf{G}raph (IRSG) to facilitate the preservation of the details in generated sketches as depicted in Fig. \ref{fig:graph}. Since different face components contain widely varying amounts of pixels, details in small components such as the two eyes might be easily overlooked. Thus, to balance the contributions of different components, we treat each component as a graph node extracted in an adaptive re-weighting pattern.

\subsubsection{Intra-class Semantic Graph}

As illustrated in Fig. \ref{fig:graph}, the synthesized sketch is represented as $F\in R^{h_{f}\times w_{f}\times c_{f}}$, where $h_{f}$, $w_{f}$ and $c_{f}$ denote the height, width, and channel of the sketch, respectively. Here, we divide the synthesized sketch into 12 facial components guided by the aforementioned semantic layouts. Thus, each node of IASG is formulated as: 
\begin{align}
\mu\begin{pmatrix}
c
\end{pmatrix} = \frac{1}{\begin{vmatrix}
S\begin{pmatrix}
:,:,c
\end{pmatrix}
\end{vmatrix}}\sum_{i=1}^{h_{f}}\sum_{j=1}^{w_{f}}S\begin{pmatrix}
i,j,c
\end{pmatrix}F\begin{pmatrix}
i,j
\end{pmatrix},
\end{align}
\begin{align}
\nu\begin{pmatrix}
c
\end{pmatrix} = \frac{1}{\begin{vmatrix}
S\begin{pmatrix}
:,:,c
\end{pmatrix}
\end{vmatrix}}\sum_{i=1}^{h_{f}}\sum_{j=1}^{w_{f}}\left \{ S\begin{pmatrix}
i,j,c
\end{pmatrix}F\begin{pmatrix}
i,j
\end{pmatrix}-\mu \begin{pmatrix}
c
\end{pmatrix} \right \}^{2},
\end{align}
where ${\begin{vmatrix}S\begin{pmatrix}:,:,c\end{pmatrix}\end{vmatrix}}$ represents the summation of pixel numbers in each facial component with the same semantic class $c$. Obviously, this strategy leverages the pixel number summation to normalize the contribution of each node, adaptive to different face components. Moreover, $\mu(c)$ is considered as the mean center of all pixels in the $\emph{c-th}$ semantic category. Furthermore, the modulation variance $\nu(c)$ can faithfully react to the semantic variation of the intra-class feature distribution. Note that both $\mu$ and $\nu$ are tensors in practice. As depicted in Fig. \ref{fig:graph}, each circle in the Graph Nodes represents different semantic classes by different colors. To model the intra-class correlation in each facial component, we calculate the cosine similarity between the synthesized sketch and each mean center (and variance center). Formally, the computation is listed as follows:
\begin{align}
\mathbb{C}_{1}=&\frac{F\cdot \mu }{\left \| F \right \|_{2}\cdot \left \| \mu  \right \|_{2}}
\notag
\\\mathbb{C}_{2}=&\frac{F\cdot \nu }{\left \| F \right \|_{2}\cdot \left \| \nu  \right \|_{2}}.
\end{align}
In this way, we construct the intra-class semantic graph for both the  synthesized sketch and the target ones.


\subsubsection{Inter-class Structure Graph}
Subsequently, to ensure the generated sketch is structure-coordinated with the target, we model the inter-class structure relations among every two facial components in IRSG. Formally, the IRSG is expressed as:
\begin{align}
\mathbb{G}_{1}= \langle \mu , \mathbb{E}_{1} \rangle
\notag
\\\mathbb{G}_{2}= \langle \nu , \mathbb{E}_{2} \rangle
\end{align}
where the $\mu$ and $\nu$ represent nodes in the structure graph. Then, we utilize the edge $\mathbb{E}$ between every two nodes to represent the structure relation of different facial components. Concretely, the graph edges are computed as the Euclidean Distance:
\begin{align}
\mathbb{E}(c_{1}, c_{2})=\mathcal{E}(\mu (c_{1}), \mu (c_{2}))
\notag
\\\mathbb{E}(c_{1}, c_{2})=\mathcal{E}(\nu (c_{1}), \nu (c_{2}))
\end{align}
where the $c \in \begin{bmatrix}1,2,\cdots,12\end{bmatrix}$ denotes different facial components and $\mathcal{E}$ represents the Euclidean Distance. Once we obtain the ITSG of both generated sketches and target ones, we exploit the target ITSG to constrain the synthesized sketches.


\begin{figure}
\begin{center}
\includegraphics[width=1\linewidth]{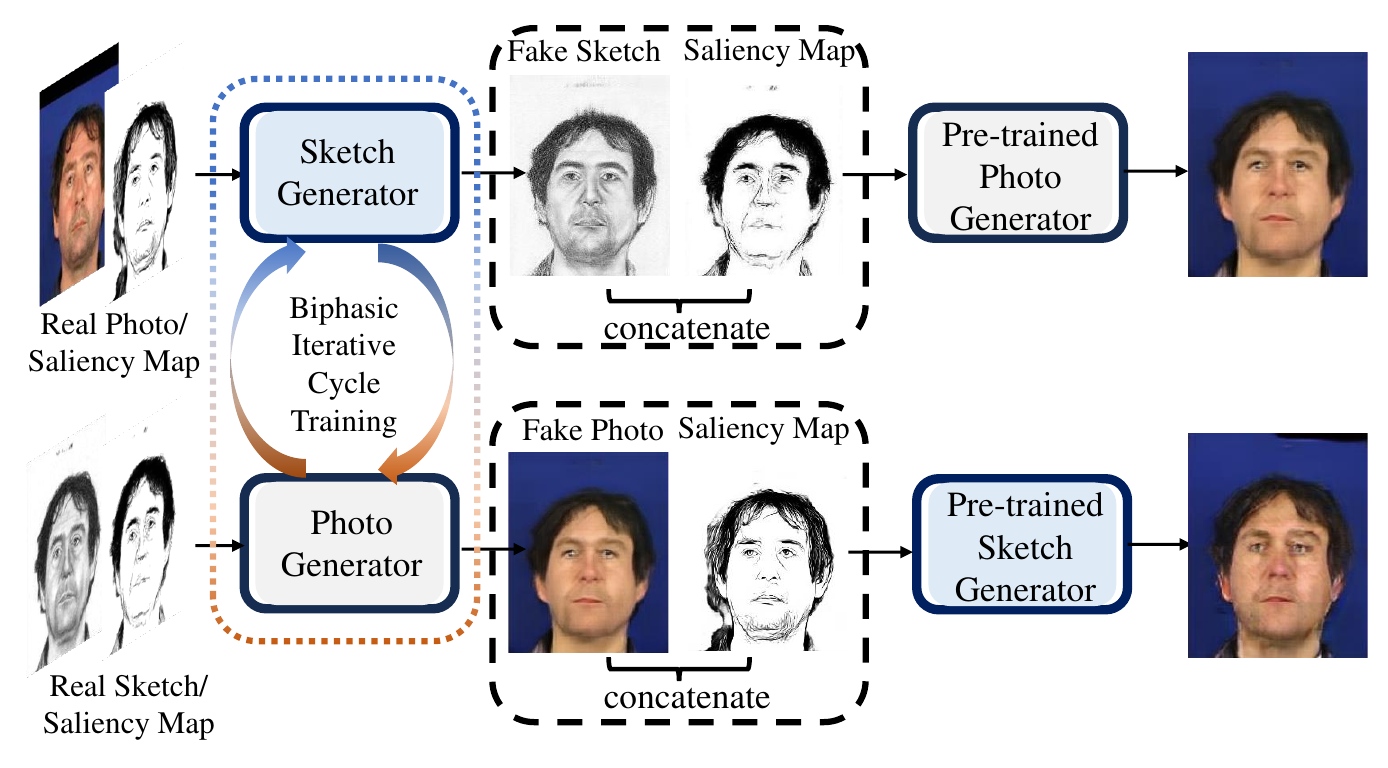}
\end{center}
   \caption{The pipeline of our proposed iterative Training Strategy. The pre-trained photo generator is leveraged as a knowledge extractor to improve the perceptual quality of the generated fake sketch, iteratively.}
\label{fig:iterative_strategy}
\end{figure}

\subsection{Iterative Training Strategy}
Inspired by the observation that paired photo-sketch shared personal characters, we propose a biphasic iterative cycle training strategy to improve the perceptual quality of the synthesized results as illustrated in Fig. \ref{fig:iterative_strategy}.
There are multi-stage in our iterative training strategy as reported in Algorithm \ref{algorithm1}. We name the generator and discriminator as $G_{k}^{i}$, $D_{k}^{i}$, $G_{o}^{i}$ and $D_{o}^{i}$, where $i\in \left \{ 0,1,\cdots ,n \right \}$ denote the iterative stages; $k$, $o$ represent sketch synthesis and photo synthesis, respectively. 

First, we train the generator and discriminator for the sketch synthesis task. Then, the photo generator and discriminator are trained similarly.
Afterward, to improve the perceptual quality of generated sketch, we take the pre-trained photo generator as a knowledge extractor to boost the sketch generator and discriminator. 
Concretely, we take the generated fake sketch concatenated with saliency detection maps (produced from real sketch) feed into the pre-trained photo generator to acquire the fake photo. Then, we exploit the corresponding real sketch concatenated with saliency map feed into the photo generator to acquire the reconstructed photo. Then we leverage the reconstructed photo and corresponding feature maps to provide multi-scale supervision of the generated fake ones, thus significantly improving the sketch generator and discriminator. Once we obtain the updated sketch generation, we utilize it to enhance the performance of the photo generator and discriminator, iteratively. Note that this iterative cycle training strategy could be conducted on multi-stage until we obtain the optimal models.



\begin{algorithm}
\small{
\caption{The iterative training strategy of sketch synthesis}
\label{algorithm1}
\KwIn{

\qquad Input photo: $X$; saliency detection map: $M_{o}$; parsing mask: $S_{o}$;

\qquad Target sketch: $Y$; saliency detection map: $M_{k}$; parsing mask: $S_{k}$; 

\qquad The number of iteration: $i$; max iteration: $T=4$;

}
\KwOut{

\qquad The optimal models $G_{k}^{i}$ and $D_{k}^{i}$;
}
\textbf{Step1:} Initialize training $G_{k}^{0}$, $D_{k}^{0}$, $G_{o}^{0}$ and $D_{o}^{0}$;

\textbf{Step2:} Iterative training;

\For{$i = 0$ to T}
{
Feed the $X$, $M_{o}$ and $S_{o}$ to train the $G_{k}^{i+1}$ and $D_{k}^{i+1}$ which obtain fake sketch $\widehat{Y}$;

Feed the $\widehat{Y}$, $M_{k}$ and $S_{k}$ to the pre-trained $G_{o}^{i}$ to obtain corresponding $\widehat{X}$; Extract multi-level feature maps from $G_{o}^{i}$ as $\widehat{F_{o}}$; 

Feed the $Y$, $M_{k}$ and $S_{k}$ to the pre-trained $G_{o}^{i}$ to obtain corresponding $\widetilde{X}$; Extract multi-level feature maps from $G_{o}^{i}$ as $\widetilde{F_{o}}$;

Compute the loss between $\widehat{X}$ and $\widetilde{X}$,  $\widehat{F_{o}}$ and $\widetilde{F_{o}}$; Back-propagate the gradients;
}
}
\end{algorithm}

\subsection{Objective Function}
The overall objective of our model includes following functions: $\mathcal{L}_{GAN}$, $\mathcal{L}_{content}$, $\mathcal{L}_{IASG}$, $\mathcal{L}_{perceptual}$, $\mathcal{L}_{BCE}$, $\mathcal{L}_{IRSG}$ and $\mathcal{L}_{ICT}$. Here, we adopt the sketch synthesis task as the prototype to elaborate the objective functions.

\subsubsection{Adversarial Loss.} The adversarial loss is leveraged to correctly distinguish the real sketches or generated  sketches. Similar to \cite{b32}, the adversarial loss is formulated as:
\begin{align}
\mathcal{L}_{GAN}&=\mathit{E}_{X,M,Y}\left [ \log D\left ( X,M,Y \right ) \right ]
\notag
\\&+{E}_{X,M}\left [ \log \left ( 1-D\left ( X,M,G\left ( X,M \right ) \right ) \right ) \right ],
\end{align}
where $X$, $Y$ and $M$ denote the source photos, target sketches and saliency detection maps.

\subsubsection{Content Loss.} In addition, we utilize the normalized $L_{1}$ distance to represent content loss.
\begin{align}
\mathcal{L}_{content}\left ( G \right )=\mathit{E}_{X,M,Y}\left [ \left \| Y-G\left ( X,M \right ) \right \|_{1} \right ].
\end{align}

\subsubsection{Perceptual Loss.}  In order to ensure the generated sketch and the target sketch maintain similar specificity, we employ the pre-trained VGG-19 net \cite{b49} as a feature extractor to obtain high-level feature representations. We compare the features after the pool1 and pool2 layers.
\begin{align}
\mathcal{L}_{perceptual}=\sum_{l=1}^{2}\left \| \omega^{l} \left ( Y \right ) -\omega^{l} \left ( G\left ( X,M \right ) \right )\right \|_{2}^{2},
\end{align}
where $\omega ^{l}\left ( \cdot  \right )$ represents the output feature maps and $l$ denotes the selected pool1 and pool2 layers.

\subsubsection{Binary Cross-Entropy Parsing Loss.} Moreover, we introduce the Binary Cross-Entropy (BCE) loss to further refine the synthesized sketch at the semantic level. We contrast the semantic mask of the synthesized sketch and the target sketch produced by the pre-trained parsing network \cite{b42}.
\begin{align}
\mathcal{L}_{BCE}=\left ( \mathbb{P}\left ( Y \right ), \mathbb{P}\left ( G\left ( X,M \right ) \right )\right ),
\end{align}
where $\mathbb{P}$ denotes the inference process of parsing network.

\subsubsection{Intra-class Semantic Graph Loss.} In practice, we extract the intra-class semantic graphs from the target sketch and synthesized sketch, respectively. Then, we reinforce the supervision of these intra-class semantic graphs to restrain the generated sketch from matching the feature distribution of the target domain. The IntrA-class Semantic Graph (IASG) loss is formulated:
\begin{align}
\mathcal{L}_{IASG}\left ( \mathbb{C}^{target}, \mathbb{C}^{fake} \right )&=\\
\notag
&\sum_{r=1}^{2}\sum_{c=1}^{12}\left \| \mathbb{C}_{r}^{target}\left ( c \right ) -\mathbb{C}_{r}^{fake}\left ( c \right )\right \|_{2}^{2},
\end{align}
where the $r \in \begin{bmatrix}1,2\end{bmatrix}$ denotes two types of nodes of IASG represented by mean center and variance center.

\begin{table*}
\centering
\caption{Results on face sketch synthesis in the CUFS and the CUFSF datasets. $\uparrow$ indicates the higher is better, $\downarrow$ indicates the lower is better. Our method reaches the \textbf{optimal} and \emph{\textbf{sub-optimal}} results in the CUFS dataset and the CUFSF dataset.}
\label{tab:sketch_synthesis}
\resizebox{\textwidth}{!}{%
\renewcommand{\arraystretch}{1.2} 
\begin{tabular}{c|c|cccccccccc|cc}
\hline
\multicolumn{2}{c|}{Model}                    & CycleGAN & Pix2Pix & MDAL  & KT   & Col-cGAN & KD+   & MSG-SARL     & SCAGAN & DP-GAN & DIR-MFP & \textbf{SDGAN } & \textbf{ours }                    \\ \hline
\multirow{6}{*}{CUFS}  & LPIPS(alex)$\downarrow$ & 0.2776   & 0.1654  & -      & 0.2297 & -        & 0.1971 & -               & -      & -      & -     & 0.1444  & \textbf{0.1432}          \\
                       & LPIPS(squeeze)$\downarrow$       & 0.1863   & 0.1156  & -      & 0.1688 & -        & 0.1471 & -               & -      & -      & -       & 0.1017 & \textbf{0.0986}          \\
                       & LPIPS(vgg-16)$\downarrow$        & 0.3815   & 0.3059  & -      & 0.3483 & -        & 0.3052 & -               & -      & -      & -       & 0.2767 & \textbf{0.2646}          \\
                       & FSIM$\uparrow$                 & 0.6829   & 0.7356  & 0.7275 & 0.7373 & -        & 0.7350 & \textbf{0.7594} & 0.716  & 0.7345 & 0.7378  & 0.7446 & \textit{\textbf{0.7494}} \\
                       & SSIM$\uparrow$                & 0.4638   & 0.5172  & 0.5280 & -      & 0.5244   & -      & 0.5288          & -      & -    & \textbf{0.5703}  & 0.5360 & \textit{\textbf{0.5493}} \\
                       & FID$\downarrow$                 & 58.394   & 44.272  & -      & -      & -        & -      & 46.39           & 34.2   & -      & -       & 33.408 &  \textbf{33.256}          \\ \hline
\multirow{6}{*}{CUFSF} & LPIPS(alex)$\downarrow$          & 0.2234   & 0.1932  & -      & 0.2522 & -        & 0.2368 & -               & -      & -      & -   &   0.1906  & \textbf{0.1867}          \\
                       & LPIPS(squeeze)$\downarrow$       & 0.1617   & 0.1422  & -      & 0.1740 & -        & 0.1619 & -               & -      & -      & -       & 0.1370 & \textbf{0.1341}          \\
                       & LPIPS(vgg-16)$\downarrow$          & 0.3787   & 0.3551  & -      & 0.3743 & -        & 0.3550 & -               & -      & -      & -       & 0.3358 & \textbf{0.3341}          \\
                       & FSIM $\uparrow$               & 0.7011   & 0.7284  & 0.7076 & 0.7311 & -        & 0.7171 & 0.7316          & 0.729  & 0.7080 & 0.7200  & 0.7328 & \textbf{0.7332}          \\
                       & SSIM$\uparrow$                & 0.3753   & 0.4204  & 0.3818 & -      & 0.4224   & -      & 0.4230          & -      & -      & 0.4215  & 0.4339 & \textbf{0.4407}          \\
                       & FID$\downarrow$                & 31.262   & 30.984  & -      & -      & -        & -      & 38.25           & \textbf{18.2}   & -      & -       &  30.594 & \textit{\textbf{24.577}} \\ \hline
\end{tabular}%
}
\end{table*}

\subsubsection{Inter-class Structure Graph Loss.} Based on our preliminary work \cite{b22}, we employ the InteR-class Structure Graphs (IRSG) to facilitate the coordinate structure preservation of synthesized sketches. 
Consequently, we apply the constraints between the IRSG of synthesized sketches and target sketches named IRSG Loss formulated as:
\begin{align}
\mathcal{L}_{IRSG}\left ( \mathbb{G}^{target}, \mathbb{G}^{fake} \right )&=\\
\notag
&\sum_{r=1}^{2}\sum_{c=1}^{12}\left \| \mathbb{G}_{r}^{target}\left ( c \right ) -\mathbb{G}_{r}^{fake}\left ( c \right )\right \|_{2}^{2}.
\end{align}
where the $r \in \begin{bmatrix}1,2\end{bmatrix}$ denotes two types of inter-class structure graphs with mean center and variance center.

\subsubsection{Iterative Cycle Training Loss.} Finally, we design the Iterative Cycle Training (ICT) loss for the biphasic iterative training strategy. Four iterations are conducted in the experiments to reach the optimal results. 
\begin{align}
\mathcal{L}_{ICT}=\sum_{l=1}^{5}\left \| G_{o}^{l,i} \left ( Y \right ) -G_{o}^{l,i} \left ( G_{k}^{i}\left ( X,M \right ) \right )\right \|_{1},
\end{align}
where $G_{o}^{l,i}\left ( \cdot  \right )$ denotes the photo generator and $i$ represents the times of iteration. We select the feature maps to extract multi-level identity-specific information represented by $l$.

\subsubsection{Full Objective.} Eventually, we combine all loss functions to achieve overall supervision:
\begin{align}
\mathcal{L}_{total} = \mathcal{L} _{GAN}&+\alpha \mathcal{L}_{content}+\lambda \mathcal{L} _{perceptual}
\notag
\\&+\delta \mathcal{L} _{BCE}+\eta \mathcal{L} _{IASG}
\notag
\\&+\tau \mathcal{L} _{IRSG}+\xi \mathcal{L} _{ICT}.
\end{align}
where the $\alpha $, $\lambda$, $\delta$, $\eta$, $\tau$ and $\xi$ are weighting factors. Furthermore, the generator $G$ and the discriminator $D$ could be optimized by the following formulation:
\begin{align}
\min_{G}\max_{D}\mathcal{L}_{total}
\end{align}

\section{Experiments}
In this section, we first introduce the implementation details of our method. Then, we describe the datasets and evaluation criteria. Next, the experimental results are presented from both quantitative and qualitative perspectives, showing the effectiveness of our proposed method. Finally, we conduct the ablation study to verify each module.

\subsection{Implementation Details}
Both the generator and discriminator are implemented on the platform Pytorch \cite{b44} with a single NVIDIA GeForce Titan X GPU. We leverage the Adam optimizer with $\beta _{1}=0.5$ and $\beta _{2}=0.999$. The total training epochs are 200, then the initial learning rate is set to 0.0002 for the first 100 epochs and decays linearly in the last 100 epochs. Additionally, we utilize the Instance Normalization \cite{b45}, and set the batchsize = 1. Meanwhile, the weighting factors are set as $\alpha=100$, $\lambda=10$, $\delta=15$, $\eta=100$, $\tau=100$ and $\xi=5$, respectively.

\begin{table*}[htbp]
\centering
\caption{Results on face photo synthesis in the CUFS and CUFSF datasets. $\uparrow$ indicates the higher is better, $\downarrow$ indicates the lower is better. Our method reaches the \textbf{optimal} and \emph{\textbf{sub-optimal}} results in the CUFS dataset and the CUFSF dataset. Note that the results of the PS2MAN model are directly borrowed from [19].}
\label{tab:photo_synthesis}
\footnotesize
\setlength{\tabcolsep}{2.5mm}{%
\renewcommand{\arraystretch}{1.2} 
\begin{tabular}{c|c|ccccccc|cc}
\hline
\multicolumn{2}{c|}{Model}              & CycleGAN & Pix2Pix & KT    & KD+    & MSG-SARL & SCAGAN         & PS2MAN  &  \textbf{SDGAN} & \textbf{ours}            \\ \hline
\multirow{6}{*}{CUFS}  & LPIPS(alex) $\downarrow$    & 0.2898   & 0.1687  & 0.1919 & 0.1717 & -        & -              & 0.2464 & 0.1674 & \textbf{0.1497}          \\
                       & LPIPS(squeeze) $\downarrow$ & 0.2509   & 0.1433  & 0.1747 & 0.1474 & -        & -              & 0.2158 & 0.1370 & \textbf{0.1225}          \\
                       & LPIPS(vgg-16) $\downarrow$  & 0.4383   & 0.3031  & 0.3208 & 0.2806 & -        & -              & 0.3254 & 0.2640 & \textbf{0.2367}          \\
                       & FSIM  $\uparrow$         & 0.7270   & 0.7723  & 0.7851 & 0.7819 & 0.7866   & 0.795          & 0.7819 & 0.7845 & \textbf{0.8001}          \\
                       & SSIM  $\uparrow$         & 0.4461   & 0.6086  & -      & -      & 0.6242   & -              & -      & 0.6543 & \textbf{0.6822}          \\
                       & FID $\downarrow$           & 124.540  & 86.996  & -      & -      & 66.17    & \textbf{40.3}  & -      & 63.937 & \textit{\textbf{49.925}} \\ \hline
\multirow{6}{*}{CUFSF} & LPIPS(alex) $\downarrow$   & 0.2271   & 0.2115  & 0.2440 & 0.2322 & -        & -              & 0.3145 & 0.2011 & \textbf{0.1998}          \\
                       & LPIPS(squeeze) $\downarrow$ & 0.1725   & 0.1669  & 0.2023 & 0.1791 & -        & -              & 0.2853 & 0.1581 & \textbf{0.1556}          \\
                       & LPIPS(vgg-16) $\downarrow$ & 0.3690   & 0.3579  & 0.3758 & 0.3565 & -        & -              & 0.4237 & 0.3422 & \textbf{0.3376}          \\
                       & FSIM  $\uparrow$         & 0.7544   & 0.7855  & 0.7931 & 0.7789 & 0.7734   & \textbf{0.845} & 0.7812 & 0.7902 & \textit{\textbf{0.7955}} \\
                       & SSIM  $\uparrow$         & 0.5594   & 0.6194  & -      & -      & 0.6114   & -              & -      & 0.6305 & \textbf{0.6441}          \\
                       & FID $\downarrow$           & 29.584   & 60.286  & -      & -      & 59.61    & \textbf{20.6}  & -      & 38.776 & \textit{\textbf{38.372}} \\ \hline
\end{tabular}%
}
\end{table*}

\begin{figure}[t]
\begin{center}
\includegraphics[width=1\linewidth]{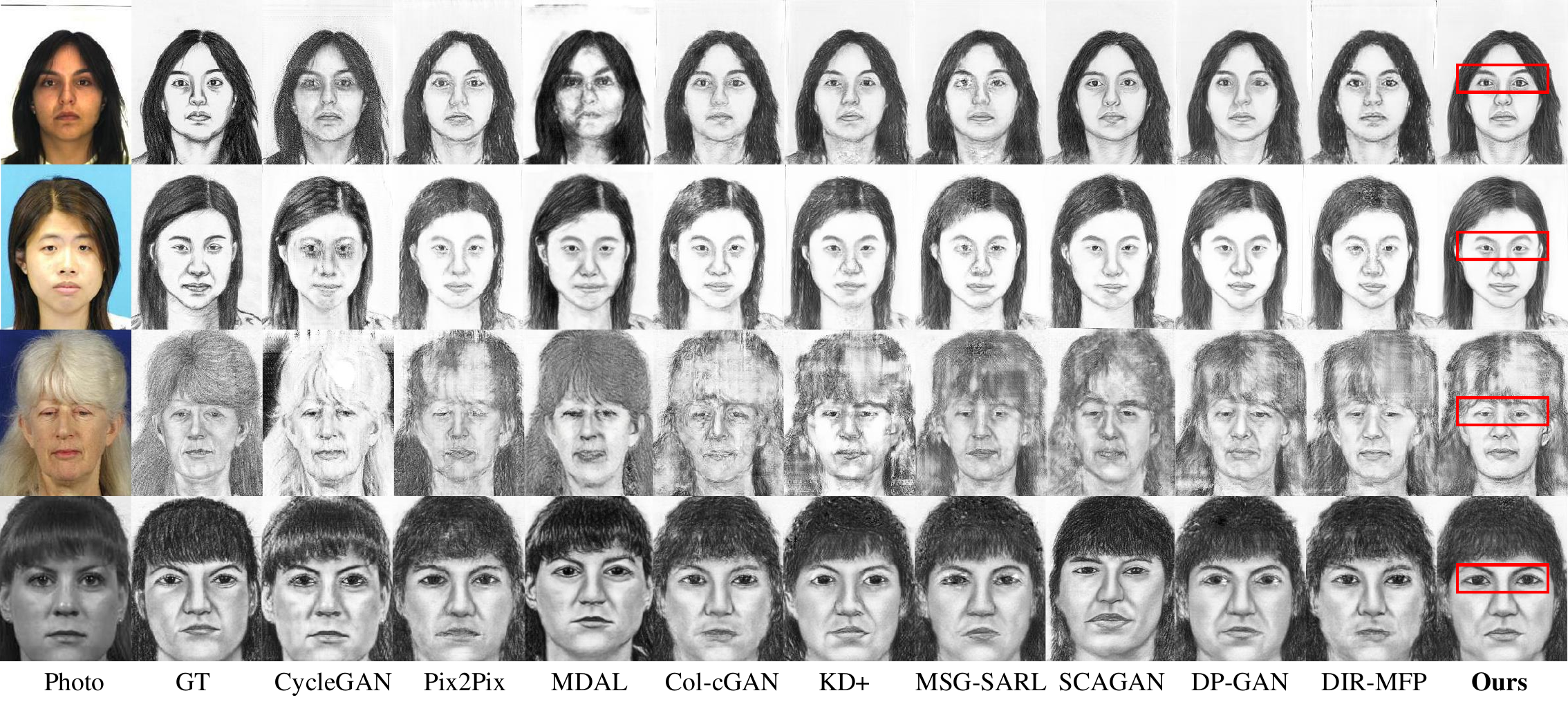}
\end{center}
   \caption{Visualization of our generated sketches against various state-of-the-art methods, \eg CycleGAN~\cite{b59}, Pix2Pix~\cite{b26}, MDAL~\cite{b16}, Col-cGAN~\cite{b61}, KD+~\cite{b19}, MSG-SARL~\cite{b18}, SCAGAN~\cite{b20}, DP-GAN~\cite{b62}, DIR-MFP\cite{b63}. From top to bottom, samples of the first three rows are from the CUFS dataset, and the others are from the CUFSF dataset. Best view on screen.} 
\label{fig:sketch_visualization}
\end{figure}

\subsection{Datasets and Evaluation Criteria}
We conduct extensive experiments on the CUHK Face Sketch Dataset (CUFS) \cite{b1} and the CUHK Face Sketch FERET Dataset (CUFSF) \cite{b46}. In the CUFS dataset, there are 606 faces, of which 188 faces are from the Chinese University of Hong Kong (CUHK) student database \cite{b3}, 123 faces from the AR database \cite{b47}, and 295 faces from the XM2VTS database \cite{b48}. For each sample, there are paired face photo-sketch drawn by the artist in natural lighting conditions. The CUFSF dataset contains 1194 face photos with paired sketches. However, all the photos in the CUFSF dataset are acquired under complex illumination variations as illustrated in Fig. \ref{fig:prior} (d). For both datasets, we exploit the geometrical alignment strategy between the photos and sketches, based on the points of two eye centers and the mouth centers. Then, the aligned photo-sketch pair is cropped to $200 \times 250$ following \cite{b49}. Meanwhile, we adopt the reshaping and padding conventions in \cite{b20} to expand the input image size to $256 \times 256$.

The experimental performance is measured by multiple metrics. We employ the Feature Similarity Index Metric (FSIM) \cite{b50} to evaluate the feature quality of synthesized sketches. FSIM is commonly utilized to measure the low-level similarity between the paired images, which extracts the phase congruence (PC) and the image gradient magnitude (GM) as features to index the quality. Consequently, blurring and noise of the generated images are evaluated by FSIM, ordinarily. 

In addition, we apply the Structural Similarity Index Metric (SSIM) \cite{b51} to demonstrate the perceptual similarity between synthesized results and ground-truth images, which follows the visibility of humans. However, some works point out that SSIM tends to favor over-smoothed images and ignores the texture of the results, which is not completely consistent with human perception \cite{b52}. Therefore, we introduce the Learned Perceptual Image Patch Similarity (LPIPS) \cite{b53} combined with SSIM to measure the perceptual visibility of synthesized results. LPIPS calculates the distance of embedding features between the generated images and target images. In this paper, LPIPS is exploited by three classification networks which are SqueezeNet \cite{b54}, AlexNet \cite{b55}, and VGGNet \cite{b43}.

Besides, we adopt the Fréchet Inception Distance (FID) \cite{b56} to compute the Earth-mover distance (EMD) of distributions between the target domain and the synthesized image domain. Specifically, a pre-trained Inception-v3 network \cite{b57} is raised to measure the 2048-dimension features between the two contrast domains. FID is widely used in biphasic photo-sketch synthesis tasks and presents high confidence in image realism.

To evaluate the preservation of personal identity characteristics of the generated human faces, we introduce the face verification rate (FVR) which is implemented by Face++ API \cite{b58}. Previously, other researchers often use Null-space Linear Discriminant Analysis (NLDA) to measure identity-specific. However, NLDA is found to be seriously affected by image texture and deformations so it might not make an accurate assessment of identity-specific characteristics \cite{b19}. We utilize the Face++ APIs which the threshold is set as 73.975@FAR=1e-5 in our identity preservation experiments.

To demonstrate the superiority of our method, we adopt several benchmark approaches (\eg, CycleGAN \cite{b59}, Pix2Pix \cite{b26}, MDAL \cite{b16}, KT \cite{b60}, Col-cGAN \cite{b61}, KD+ \cite{b19}, MSG-SARL \cite{b18}, SCAGAN \cite{b20}, DP-GAN \cite{b62}, DIP-MFP \cite{b63} and PS2MAN \cite{b64}) for comparison. In addition, to further demonstrate the effectiveness of the proposed method, we also conduct comparison experiments between our preliminary work \textbf{SDGAN} \cite{b22} and our current model.

\begin{figure}[h]
\begin{center}
\includegraphics[width=1\linewidth]{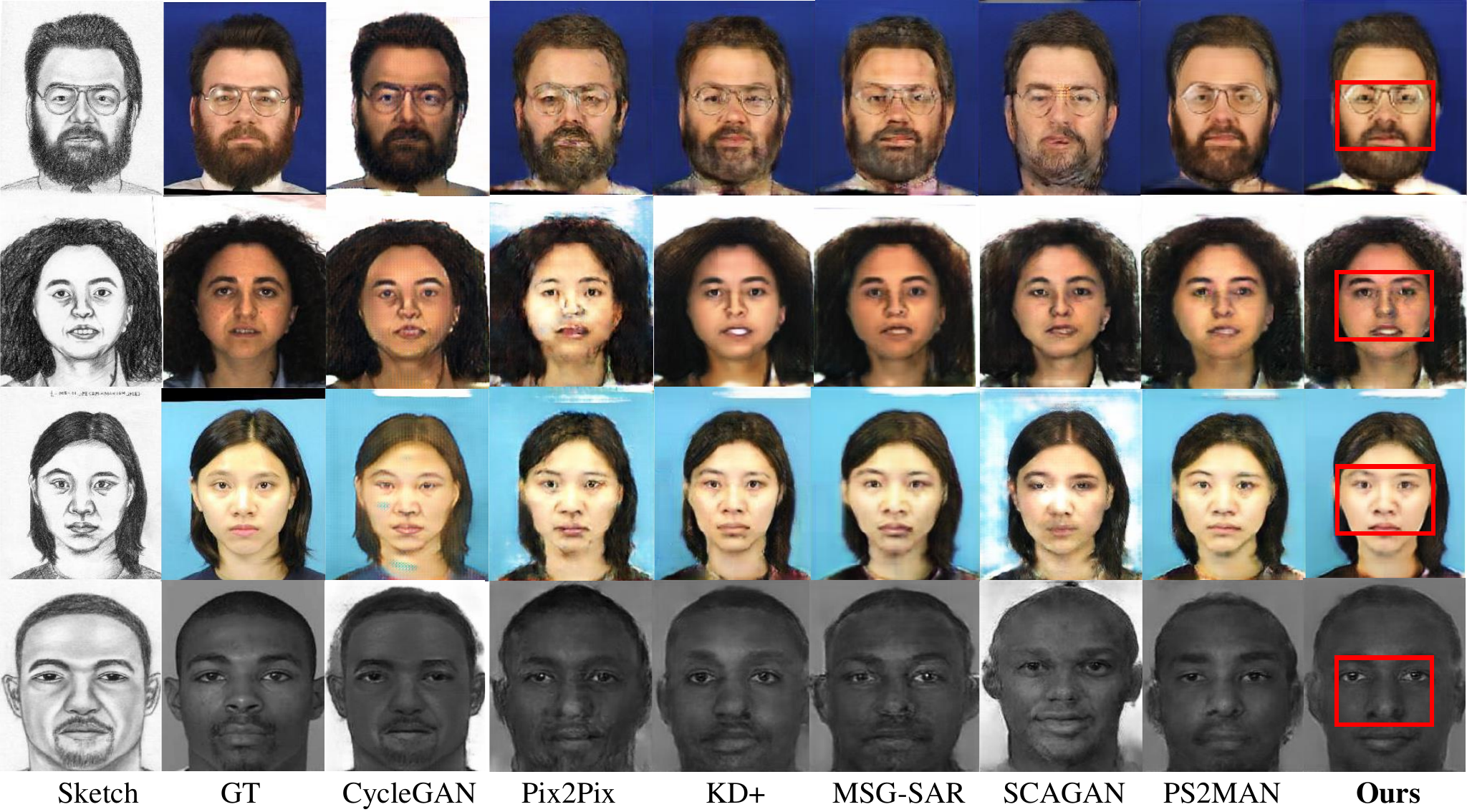}
\end{center}
   \caption{Visualization of our generated photos against various state-of-the-art methods, \eg CycleGAN~\cite{b59}, Pix2Pix~\cite{b26}, KD+~\cite{b19}, MSG-SARL~\cite{b18}, SCAGAN~\cite{b20}, PS2MAN~\cite{b64}. From top to bottom, samples of the first three rows are from the CUFS dataset, and the others are from the CUFSF dataset. Best view on screen.}
\label{fig:photo_visualization}
\end{figure}

\subsection{Results on face sketch synthesis task}
We verify the performance of our method on the face sketch synthesis on the CUFS and the CUFSF datasets. The experimental results are reported in Table \ref{tab:sketch_synthesis}. Our method achieves the best capability on the indicators of LPIPS (alex), LPIPS (squeeze), LPIPS (vgg-16), and FID in the CUFS dataset. The proposed method decreases the previous best FID from 34.2 to 33.256. In addition, our method realizes the best performance on all indicators in the CUFSF dataset except FID. Obviously, the LIPIS (alex, squeeze, vgg) is significantly decreased with the large margin in both datasets which implies the higher fidelity and realism of the sketches generated by our method. In the sketch synthesis experiments, the MSG-SARL, DIR-MFP, and SCAGAN reach better results in limited metrics on the FSIM, SSIM, and FID. We observe that in Fig.~\ref{fig:sketch_visualization}, the generated images of these methods are more smoothed. Since the metrics such as SSIM is more kind to the smoothed image texture, this leads to better results than ours. However, we exploit the saliency map to provide global structure information. This results in better performance on the LPIPS and a more structural-coordinated human face.

We visualize the comparison of generated sketches among our method and various state-of-the-art counterparts as depicted in Fig.~\ref{fig:sketch_visualization}. Since the KT~\cite{b5} is the preliminary conference version of KD+~\cite{b4}, we only display the better version in the experiments.
We observe that CycleGAN and MDAL produce low-fidelity sketches compared with ground truth. The results of Col-cGAN, MSG-SARL, and SCAGAN contain blur and noisy effects. Conversely, our method achieves the best performance, especially in the detailed facial components such as the eyes, ears, and hair contours (\eg, in the red boxes of the last column).


\subsection{Results on face photo synthesis task}
Compared with face sketch synthesis, there are fewer previous methods that focus on the photo synthesis task. Our method obtains the best performance with LPIPS (alex, squeeze, vgg) and SSIM on both datasets. According to Table \ref{tab:photo_synthesis}, the SCAGAN \cite{b20} achieves better results than our method with FSIM and FID because of its stacking architecture. In contrast, our network is end-to-end that has stronger practical significance on edge devices. It is more challenging to synthesize photos from the CUFSF dataset. As shown in Fig. \ref{fig:prior} (d), there are illumination variations in the CUFSF dataset. Meanwhile, all photos are grayscale and sketches are drawn with deformation. 

We display the visual comparison in Fig.~\ref{fig:photo_visualization}. Pix2Pix, MSG-SAR, and SCAGAN synthesize the natural photos compared with the ground truth faces. CycleGAN, KD+, and PS2MAN generate better results while their faces contain less stylized textures of the painter. On the contrary, our method achieves vivid and realistic photo synthesis. As highlighted in the red boxes, the facial structures of our results are much more natural than other competitors.

\begin{table}[t]
\centering
\caption{Comparison between the preliminary work \textbf{SDGAN} and our current network named \textbf{Ours} in the Table All the performances of our current model are \textbf{BETTER} than preliminary work. $\uparrow$ indicates the higher is better, $\downarrow$ indicates the lower is better.}
\label{tab:preliminary}
\footnotesize
\setlength{\tabcolsep}{1mm}{%
\renewcommand{\arraystretch}{1.1}
\begin{tabular}{cc|ccc|ccc}
\hline
\multicolumn{2}{c|}{Task} & \multicolumn{3}{c|}{\begin{tabular}[c]{@{}c@{}}Sketch\\ Synthesis\end{tabular}} & \multicolumn{3}{c}{\begin{tabular}[c]{@{}c@{}}Photo\\ Synthesis\end{tabular}} \\ \hline
\multicolumn{2}{c|}{Model} & SDGAN & \multicolumn{1}{c|}{\textbf{Ours}} & Gap & SDGAN & \multicolumn{1}{c|}{\textbf{Ours}} & Gap \\ \hline
\multicolumn{1}{c|}{\multirow{7}{*}{CUFS}} & \begin{tabular}[c]{@{}c@{}}LPLPS\\ (alex)$\downarrow$\end{tabular} & 0.1444 & \multicolumn{1}{c|}{\textbf{0.1432}} & 0.0012 & 0.1674 & \multicolumn{1}{c|}{\textbf{0.1497}} & 0.0177 \\
\multicolumn{1}{c|}{} & \begin{tabular}[c]{@{}c@{}}LPIPS\\ (squeeze)$\downarrow$\end{tabular} & 0.1017 & \multicolumn{1}{c|}{\textbf{0.0986}} & 0.0031 & 0.1370 & \multicolumn{1}{c|}{\textbf{0.1225}} & 0.0145 \\
\multicolumn{1}{c|}{} & \begin{tabular}[c]{@{}c@{}}LPIPS\\ (vgg-16)$\downarrow$\end{tabular} & 0.2767 & \multicolumn{1}{c|}{\textbf{0.2646}} & 0.0021 & 0.2640 & \multicolumn{1}{c|}{\textbf{0.2367}} & 0.0273 \\
\multicolumn{1}{c|}{} & FSIM $\uparrow$ & 0.7446 & \multicolumn{1}{c|}{\textbf{0.7494}} & 0.0048 & 0.7845 & \multicolumn{1}{c|}{\textbf{0.8001}} & 0.0156 \\
\multicolumn{1}{c|}{} & SSIM $\uparrow$ & 0.5360 & \multicolumn{1}{c|}{\textbf{0.5493}} & 0.0133 & 0.6543 & \multicolumn{1}{c|}{\textbf{0.6822}} & 0.0279 \\
\multicolumn{1}{c|}{} & FID $\downarrow$ & 33.408 & \multicolumn{1}{c|}{\textbf{33.256}} & 0.152 & 63.937 & \multicolumn{1}{c|}{\textbf{49.925}} & 14.012 \\
\multicolumn{1}{c|}{} & FVR $\uparrow$ & 86.179 & \multicolumn{1}{c|}{\textbf{87.542}} & 1.363 & 79.040 & \multicolumn{1}{c|}{\textbf{81.780}} & 2.74 \\ \hline
\multicolumn{1}{c|}{\multirow{7}{*}{CUFSF}} & \begin{tabular}[c]{@{}c@{}}LPIPS\\ (alex)$\downarrow$\end{tabular} & 0.1906 & \multicolumn{1}{c|}{\textbf{0.1867}} & 0.0039 & 0.2011 & \multicolumn{1}{c|}{\textbf{0.1998}} & 0.0013 \\
\multicolumn{1}{c|}{} & \begin{tabular}[c]{@{}c@{}}LPIPS\\ (squeeze)$\downarrow$\end{tabular} & 0.1370 & \multicolumn{1}{c|}{\textbf{0.1341}} & 0.0029 & 0.1581 & \multicolumn{1}{c|}{\textbf{0.1556}} & 0.0025 \\
\multicolumn{1}{c|}{} & \begin{tabular}[c]{@{}c@{}}LPIPS\\ (vgg-16)$\downarrow$\end{tabular} & 0.3358 & \multicolumn{1}{c|}{\textbf{0.3341}} & 0.0017 & 0.3422 & \multicolumn{1}{c|}{\textbf{0.3376}} & 0.0046 \\
\multicolumn{1}{c|}{} & FSIM $\uparrow$ & 0.7328 & \multicolumn{1}{c|}{\textbf{0.7332}} & 0.0004 & 0.7902 & \multicolumn{1}{c|}{\textbf{0.7955}} & 0.0053 \\
\multicolumn{1}{c|}{} & SSIM $\uparrow$ & 0.4339 & \multicolumn{1}{c|}{\textbf{0.4407}} & 0.0068 & 0.6305 & \multicolumn{1}{c|}{\textbf{0.6441}} & 0.0136 \\
\multicolumn{1}{c|}{} & FID $\downarrow$ & 30.594 & \multicolumn{1}{c|}{\textbf{24.577}} & 6.017 & 38.776 & \multicolumn{1}{c|}{\textbf{38.372}} & 0.404 \\
\multicolumn{1}{c|}{} & FVR $\uparrow$ & 86.758 & \multicolumn{1}{c|}{\textbf{87.109}} & 0.351 & 62.763 & \multicolumn{1}{c|}{\textbf{63.344}} & 0.581 \\ \hline
\end{tabular}
}
\end{table}

\subsection{Comparison with preliminary work}
To evaluate the improvement between our current model and preliminary work \textbf{SDGAN}, we conduct the performance comparison as reported in Table \ref{tab:preliminary}. 

\begin{figure}
\begin{center}
\includegraphics[width=0.97\linewidth]{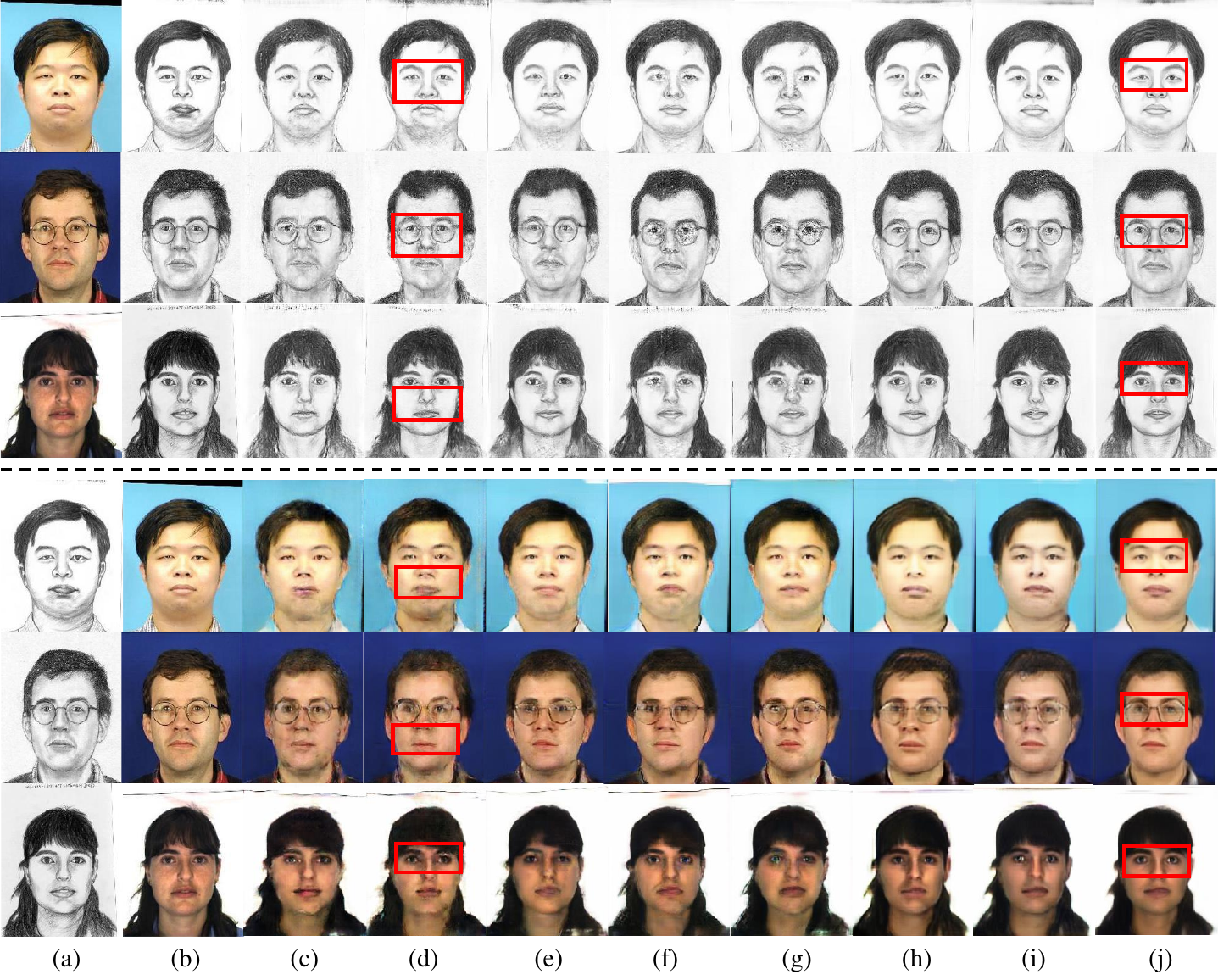}
\end{center}
   \caption{Ablation studies of synthesized sketches and photos on the CUFS dataset. From top to bottom, the examples are selected from the XM2VTS database, CUHK database, and AR database. (a) Source images, (b) Target Images, (c) Pix2Pix, (d) w/ $M$ , (e) w/ $M$ + $S$, (f) w/ $M$ + $S$ + Perceptual Loss, (g) w/ $M$ + $S$ + Perceptual Loss + BCE Loss, (h) w/ $M$ + $S$ + Perceptual Loss + BCE Loss + IASG Loss, (i) w/ $M$ + $S$ + Perceptual Loss + BCE Loss + IASG Loss + IRSG Loss, (j) w/ $M$ + $S$ + Perceptual Loss + BCE Loss + IASG Loss + IRSG Loss + ICT Loss. Note that the \emph{\textbf{(h)}} column represents the results of our preliminary work \textbf{SDGAN}.}
   \vspace{-1em}
\label{fig:CUFS}
\end{figure}

\begin{figure}
\begin{center}
\includegraphics[width=0.97\linewidth]{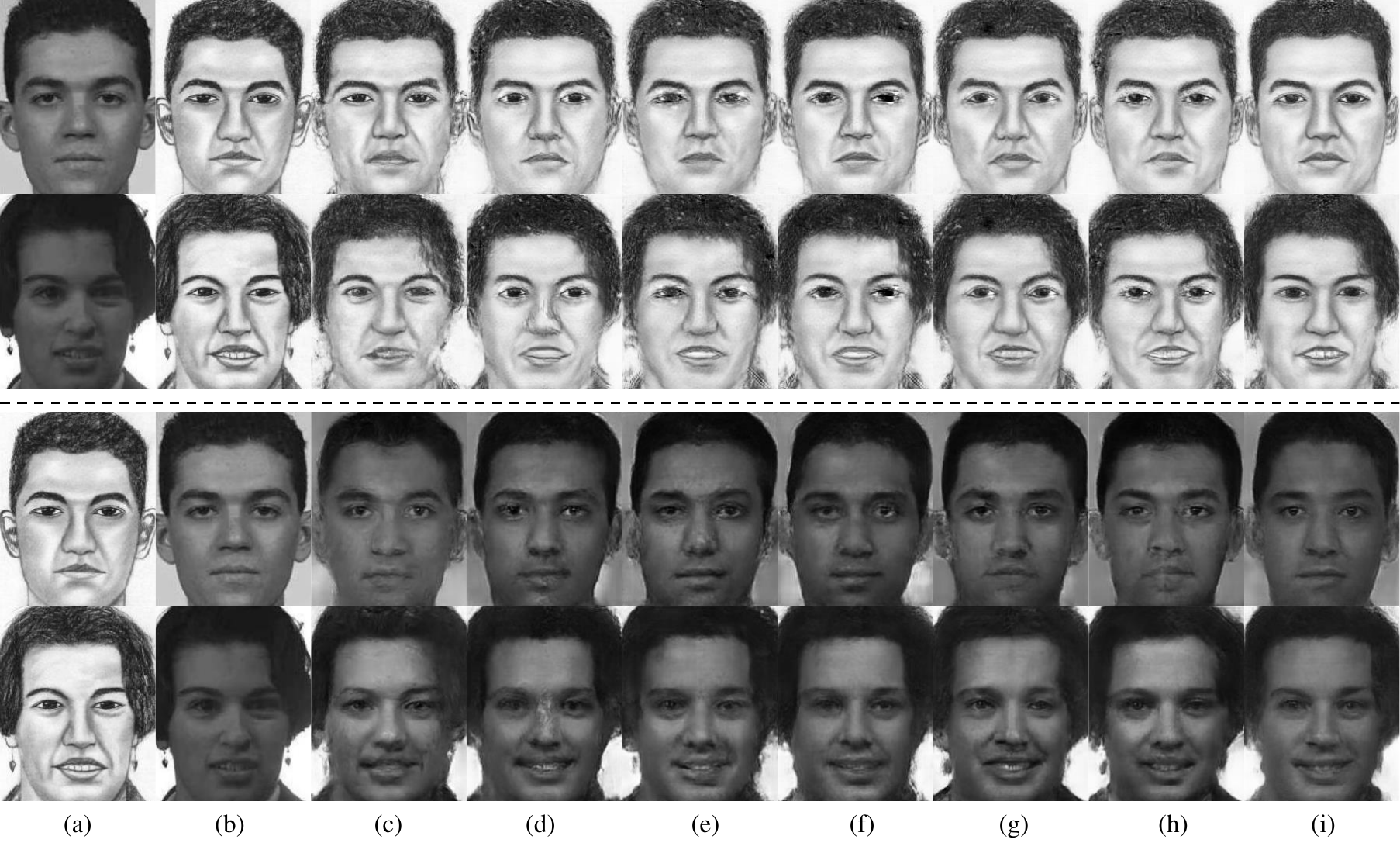}
\end{center}
   \caption{Ablation studies of synthesized sketches and photos on the CUFSF dataset. (a) Source images, (b) Target Images, (c) Pix2Pix, (d) w/ $S$, (e) w/ $S$ + Perceptual Loss, (f) w/ $S$ + Perceptual Loss + BCE Loss, (g) w/ $S$ + Perceptual Loss + BCE Loss + IAG Loss, (h) w/ $S$ + Perceptual Loss + BCE Loss + IASG Loss + IRSG Loss, (i) w/ $S$ + Perceptual Loss + BCE Loss + IASG Loss + IRSG Loss + ICT Loss. Note that the \emph{\textbf{(g)}} column represents the results of our preliminary work \textbf{SDGAN}.}
   \vspace{-1em}
\label{fig:CUFSF}
\end{figure}

Since SDGAN supervises the synthesis of sketches and photos in the intra-class semantic space, the generated human faces always have abundant detailed features. However, SDGAN does not effectively extract the structure knowledge in the facial images, which is modeled by the inter-class structure graph in the current framework. Therefore, we take advantage of both intra-class semantic and inter-class structure graph representations that allow the synthesized human faces to be more clearly structured and contain distinctive features, as illustrated in columns (h-j) of Fig. \ref{fig:CUFS} and columns (g-i) of Fig. \ref{fig:CUFSF}. Besides, the SDGAN is designed for unidirectional photo-sketch synthesis that overlooks there are personal-identity consistency between the paired photo-sketch.
On the contrary, with the help of the biphasic iterative cycle training strategy, the identity-preserving ability of our current model is significantly enhanced. This is essential in the image-to-image translation task with geometrical alignment datasets. 
As reported in Table \ref{tab:ablation-CUFS} and Table \ref{tab:ablation_CUFSF}, the Face verification Rate (FVR) significantly increased after applying the biphasic iterative cycle training strategy. Besides, as illustrated in Table \ref{tab:preliminary}, our current model performance has been significantly raised compared with preliminary \textbf{SDGAN}, especially in the photo synthesis task. From a more intuitive point of view, the human faces synthesized by the current model give the expression of more fidelity and consistency than \textbf{SDGAN}.

\begin{table*}[t]
\centering
\caption{The ablation study on CUFS dataset. $\uparrow$ indicates the higher is better, $\downarrow$ indicates the lower is better. Note that our preliminary work \textbf{SDGAN} is indicated as the last \textbf{\emph{3rd}} row from up to bottom in each task.}
\label{tab:ablation-CUFS}
\resizebox{\textwidth}{!}{%
\renewcommand{\arraystretch}{1.1} 
\begin{tabular}{c|cccccccc|ccccccc}
\hline
\multicolumn{9}{c|}{Model Variations} & \multicolumn{7}{c}{Criterion} \\ \hline
 & Backbone & \begin{tabular}[c]{@{}c@{}}Salicency\\ Mask $M$\end{tabular} & \begin{tabular}[c]{@{}c@{}}Parsing\\ Layouts $S$\end{tabular} & \begin{tabular}[c]{@{}c@{}}Perceptual\\ Loss\end{tabular} & \begin{tabular}[c]{@{}c@{}}BCE \\ Loss\end{tabular} & \begin{tabular}[c]{@{}c@{}}IASG \\ Loss\end{tabular} & \begin{tabular}[c]{@{}c@{}}IRSG\\ Loss\end{tabular} & \begin{tabular}[c]{@{}c@{}}ICT\\ Loss\end{tabular} & \begin{tabular}[c]{@{}c@{}}LPIPS\\ (alex)$\downarrow$\end{tabular} & \begin{tabular}[c]{@{}c@{}}LPIPS\\ (squeeeze)$\downarrow$\end{tabular} & \begin{tabular}[c]{@{}c@{}}LPIPS\\ (vgg-16)$\downarrow$\end{tabular} & FSIM$\uparrow$ & SSIM$\uparrow$ & FID$\downarrow$ & \textbf{FVR}$\uparrow$ \\ \hline
\multirow{8}{*}{\begin{tabular}[c]{@{}c@{}}Sketch\\ Synthesis\end{tabular}} & $\checkmark$ & - & - & - & - & - & - & - & 0.1654 & 0.1156 & 0.3059 & 0.7356 & 0.5172 & 44.272 & 84.252 \\
 & $\checkmark$ & $\checkmark$ & - & - & - & - & - & - & 0.1581 & 0.1141 & 0.3080 & 0.7376 & 0.5145 & 42.367 & 84.504 \\
 & $\checkmark$ & $\checkmark$ & $\checkmark$ & - & - & - & - & - & 0.1570 & 0.1108 & 0.2965 & 0.7411 & 0.5286 & 41.048 & 84.698 \\
 & $\checkmark$ & $\checkmark$ & $\checkmark$ & $\checkmark$ & - & - & - & - & 0.1644 & 0.1123 & 0.2918 & 0.7424 & 0.5248 & 34.854 & 83.893 \\
 & $\checkmark$ & $\checkmark$ & $\checkmark$ & $\checkmark$ & $\checkmark$ & - & - & - & 0.1642 & 0.1138 & 0.2918 & 0.7433 & 0.5311 & 34.638 & 84.892 \\
 & $\checkmark$ & $\checkmark$ & $\checkmark$ & $\checkmark$ & $\checkmark$ & $\checkmark$ & - & - & 0.1444 & 0.1017 & 0.2767 & 0.7446 & 0.5360 & 33.408 & 86.179 \\
 & $\checkmark$ & $\checkmark$ & $\checkmark$ & $\checkmark$ & $\checkmark$ & $\checkmark$ & $\checkmark$ & - & 0.1446 & 0.0995 & 0.2727 & 0.7466 & 0.5482 & \textbf{33.166} & 86.648 \\
 & $\checkmark$ & $\checkmark$ & $\checkmark$ & $\checkmark$ & $\checkmark$ & $\checkmark$ & $\checkmark$ & $\checkmark$ & \textbf{0.1432} & \textbf{0.0986} & \textbf{0.2646} & \textbf{0.7494} & \textbf{0.5493} & \textit{\textbf{33.256}} & \textbf{87.542} \\ \hline
\multirow{8}{*}{\begin{tabular}[c]{@{}c@{}}Photo\\ Synthesis\end{tabular}} & $\checkmark$ & - & - & - & - & - & - & - & 0.1687 & 0.1433 & 0.3031 & 0.7723 & 0.6086 & 86.996 & 73.527 \\
 & $\checkmark$ & $\checkmark$ & - & - & - & - & - & - & 0.1660 & 0.1449 & 0.3018 & 0.7742 & 0.6093 & 93.204 & 74.046 \\
 & $\checkmark$ & $\checkmark$ & $\checkmark$ & - & - & - & - & - & 0.1589 & 0.1341 & 0.2788 & 0.7759 & 0.6266 & 66.176 &  75.260 \\
 & $\checkmark$ & $\checkmark$ & $\checkmark$ & $\checkmark$ & - & - & - & - & 0.1609 & 0.1341 & 0.2799 & 0.7777 & 0.6303 & 64.495 & 76.427 \\
 & $\checkmark$ & $\checkmark$ & $\checkmark$ & $\checkmark$ & $\checkmark$ & - & - & - & 0.1562 & 0.1326 & 0.2760 & 0.7788 & 0.6320 & 65.729 & 76.778 \\
 & $\checkmark$ & $\checkmark$ & $\checkmark$ & $\checkmark$ & $\checkmark$ & $\checkmark$ & - & - & 0.1674 & 0.1370 & 0.2640 & 0.7845 & 0.6543 & 63.937 & 79.040 \\
 & $\checkmark$ & $\checkmark$ & $\checkmark$ & $\checkmark$ & $\checkmark$ & $\checkmark$ & $\checkmark$ & - & 0.1571 & 0.1306 & 0.2588 & 0.7892 & 0.6627 & 61.726 & 78.438 \\
 & $\checkmark$ & $\checkmark$ & $\checkmark$ & $\checkmark$ & $\checkmark$ & $\checkmark$ & $\checkmark$ & $\checkmark$ & \textbf{0.1497} & \textbf{0.1225} & \textbf{0.2367} & \textbf{0.8001} & \textbf{0.6822} & \textbf{49.925} & \textbf{81.780} \\ \hline
\end{tabular}%
}
\end{table*}

\begin{table*}[htbp]
\centering
\caption{The ablation study on CUFSF dataset. $\uparrow$ indicates the higher is better, $\downarrow$ indicates the lower is better. Note that our preliminary work \textbf{SDGAN} is indicated as the last \textbf{\emph{3rd}} row from up to bottom in each task.}
\label{tab:ablation_CUFSF}
\resizebox{\textwidth}{!}{%
\renewcommand{\arraystretch}{1.1} 
\begin{tabular}{c|cccccccc|ccccccc}
\hline
\multicolumn{9}{c|}{Model Variations} & \multicolumn{7}{c}{Criterion} \\ \hline
 & Backbone & \begin{tabular}[c]{@{}c@{}}Salicency\\ Mask $M$\end{tabular} & \begin{tabular}[c]{@{}c@{}}Parsing\\ Layouts $S$\end{tabular} & \begin{tabular}[c]{@{}c@{}}Perceptual\\ Loss\end{tabular} & \begin{tabular}[c]{@{}c@{}}BCE \\ Loss\end{tabular} & \begin{tabular}[c]{@{}c@{}}IASG \\ Loss\end{tabular} & \begin{tabular}[c]{@{}c@{}}IRSG\\ Loss\end{tabular} & \begin{tabular}[c]{@{}c@{}}ICT\\ Loss\end{tabular} & \begin{tabular}[c]{@{}c@{}}LPIPS\\ (alex)$\downarrow$\end{tabular} & \begin{tabular}[c]{@{}c@{}}LPIPS\\ (squeeeze)$\downarrow$\end{tabular} & \begin{tabular}[c]{@{}c@{}}LPIPS\\ (vgg-16)$\downarrow$\end{tabular} & FSIM$\uparrow$ & SSIM$\uparrow$ & FID$\downarrow$ & \textbf{FVR}$\uparrow$ \\ \hline
\multirow{8}{*}{\begin{tabular}[c]{@{}c@{}}Sketch\\ Synthesis\end{tabular}} & $\checkmark$ & - & - & - & - & - & - & - & 0.1932 & 0.1422 & 0.3551 & 0.7284 & 0.4204 & 30.984 & 86.027 \\
 & $\checkmark$ & $\checkmark$ & - & - & - & - & - & - & 0.1939 & 0.1429 & 0.3562 & 0.7275 & 0.4193 & 29.765 & 85.772 \\
 & $\checkmark$ & - & $\checkmark$ & - & - & - & - & - & 0.2037 & 0.1443 & 0.3454 & 0.7290 & 0.4297 & 30.970 & 86.052 \\
 & $\checkmark$ & - & $\checkmark$ & $\checkmark$ & - & - & - & - & 0.2070 & 0.1423 & 0.3545 & 0.7299 & 0.4270 & 28.639 & 86.099 \\
 & $\checkmark$ & - & $\checkmark$ & $\checkmark$ & $\checkmark$ & - & - & - & 0.1983 & 0.1420 & 0.3460 & 0.7300 & 0.4282 & 25.723 & 86.512 \\
 & $\checkmark$ & - & $\checkmark$ & $\checkmark$ & $\checkmark$ & $\checkmark$ & - & - & 0.1906 & 0.1370 & 0.3358 & 0.7328 & 0.4339 & 30.594 & 86.758 \\
 & $\checkmark$ & - & $\checkmark$ & $\checkmark$ & $\checkmark$ & $\checkmark$ & $\checkmark$ & - & 0.1896 & 0.1370 & 0.3349 & 0.7321 & 0.4355 & 28.995 & 87.023 \\
 & $\checkmark$ & - & $\checkmark$ & $\checkmark$ & $\checkmark$ & $\checkmark$ & $\checkmark$ & $\checkmark$ & \textbf{0.1867} & \textbf{0.1341} & \textbf{0.3341} & \textbf{0.7332} & \textbf{0.4407} & \textbf{24.577} & \textbf{87.109}  \\ \hline
\multirow{7}{*}{\begin{tabular}[c]{@{}c@{}}Photo\\ Synthesis\end{tabular}} & $\checkmark$ & - & - & - & - & - & - & - & 0.2115 & 0.1669 & 0.3579 & 0.7855 & 0.6194 & 60.286 & 62.074 \\
 & $\checkmark$ & - & $\checkmark$ & - & - & - & - & - & 0.2137 & 0.1653 & 0.3546 & 0.7875 & 0.6243 & 42.517 & 61.322 \\
 & $\checkmark$ & - & $\checkmark$ & $\checkmark$ & - & - & - & - & 0.2106 & 0.1672 & 0.3537 & 0.7882 & 0.6219 & 51.397 & 62.174 \\
 & $\checkmark$ & - & $\checkmark$ & $\checkmark$ & $\checkmark$ & - & - & - & 0.2091 & 0.1650 & 0.3497 & 0.7890 & 0.6269 & 49.787 & 62.026 \\
 & $\checkmark$ & - & $\checkmark$ & $\checkmark$ & $\checkmark$ & $\checkmark$ & - & - & 0.2011 & 0.1581 & 0.3422 & 0.7902 & 0.6305 & 38.776 & 62.763 \\
 & $\checkmark$ & - & $\checkmark$ & $\checkmark$ & $\checkmark$ & $\checkmark$ & $\checkmark$ & - & 0.2030 & 0.1593 & 0.3403 & 0.7929 & 0.6426 & \textbf{37.352} & 63.052 \\
 & $\checkmark$ & - & $\checkmark$ & $\checkmark$ & $\checkmark$ & $\checkmark$ & $\checkmark$ & $\checkmark$ & \textbf{0.1998} & \textbf{0.1556} & \textbf{0.3376} & \textbf{0.7955} & \textbf{0.6441} & \textit{\textbf{38.372}} & \textbf{63.344}  \\ \hline
\end{tabular}%
}
\end{table*}

\begin{table*}[htbp]
\centering
\caption{Ablation Study of Biphsic Iterative Training Strategy. $\uparrow$ indicates the higher is better, $\downarrow$ indicates the lower is better. }
\label{tab:Iteration_Training}
\resizebox{\textwidth}{!}{%
\renewcommand{\arraystretch}{1.05} 
\begin{tabular}{c|c|ccccc|ccccc}
\hline
\multicolumn{2}{c|}{} & \multicolumn{5}{c|}{Sketch Synthesis} & \multicolumn{5}{c}{Photo Synthesis} \\ \hline
\multicolumn{2}{c|}{Iteration} & Initial-0 & Iteration-1 & Iteration-2 & Iteration-3 & Iteration-4 & Initial-0 & Iteration-1 & Iteration-2 & Iteration-3 & Iteration-4 \\ \hline
\multirow{7}{*}{CUFS} & LPIPS(alex) $\downarrow$ & 0.1446 & 0.1447 & \textbf{0.1406} & 0.1432 & 0.1483 & 0.1571 & 0.1497 & 0.1563 & \textbf{0.1497} & 0.1518 \\
 & LPIPS(squeeze) $\downarrow$ & 0.0995 & 0.0991 & 0.0988 & \textbf{0.0986} & 0.1007 & 0.1306 & 0.1237 & 0.1306 & \textbf{0.1225} & 0.1248 \\
 & LPIPS(vgg-16) $\downarrow$ & 0.2727 & 0.2665 & 0.2650 & \textbf{0.2646} & 0.2656 & 0.2588 & 0.2449 & 0.2442 & \textbf{0.2367} & 0.2392 \\
 & FSIM $\uparrow$ & 0.7466 & 0.7468 & 0.7482 & \textbf{0.7494} & 0.7474 & 0.7892 & 0.7918 & 0.7973 & \textbf{0.8001} & 0.7980 \\
 & SSIM $\uparrow$ & 0.5482 & 0.5419 & 0.5472 & \textbf{0.5493} & 0.5424 & 0.6627 & 0.6704 & 0.6778 & \textbf{0.6822} & 0.6767 \\
 & FID $\downarrow$ & 33.166 & 30.640 & \textbf{30.092} & 33.256 & 34.490 & 61.726 & 55.804 & 54.303 & \textbf{49.925} & 51.149 \\
 & FVR $\uparrow$ & 86.648 & 87.382 & 87.523 & \textbf{87.542} & 87.359 & 78.438 & 79.922 & 80.423 & \textbf{81.790} & 80.249 \\ \hline
\multirow{7}{*}{CUFSF} & LPIPS(alex) $\downarrow$ & 0.1896 & 0.1883 & 0.1904 & \textbf{0.1867} & 0.1912 & 0.2030 & 0.2040 & 0.2038 & \textbf{0.1998} & 0.2011 \\
 & LPIPS(squeeze) $\downarrow$ & 0.1370 & 0.1361 & 0.1351 & \textbf{0.1341} & 0.1352 & 0.1593 & 0.1593 & 0.1585 & \textbf{0.1556} & 0.1565 \\
 & LPIPS(vgg-16) $\downarrow$ & 0.3349 & 0.3372 & 0.3347 & \textbf{0.3341} & 0.3377 & 0.3403 & 0.3424 & 0.3419 & \textbf{0.3376} & 0.3393 \\
 & FSIM $\uparrow$ & 0.7321 & 0.7323 & 0.7322 & \textbf{0.7332} & 0.7330 & 0.7929 & 0.7940 & 0.7941 & \textbf{0.7955} & 0.7947 \\
 & SSIM $\uparrow$ & 0.4355 & 0.4371 & 0.4384 & \textbf{0.4407} & 0.4375 & 0.6426 & 0.6410 & 0.6430 & \textbf{0.6441} & 0.6411 \\
 & FID $\downarrow$ & 28.995 & 28.462 & 25.690 & \textbf{24.577} & 26.031 & \textbf{37.352} & 46.418 & 43.447 & 38.372 & 40.534 \\
 & FVR $\uparrow$ & 87.023 & 87.099 & 87.041 & \textbf{87.109} & 86.980 & 63.052 & 63.228 & 63.191 & \textbf{63.344} & 62.774 \\ \hline
\end{tabular}%
}
\end{table*}

\subsection{Ablation Study}
In this section, we conduct extensive ablation experiments to verify the effectiveness of each module and the loss function we proposed. 

\subsubsection{Saliency Detection Map $M$}
As reported in Table \ref{tab:ablation-CUFS}, we concatenate the saliency detection map $M$ with input images to provide overall structure prior knowledge. We observe that the SSIM is affected after the utilization of the saliency map since SSIM is more sensitive to texture information. However, as a whole, the performance of the model is improved, especially the FSIM indicators. As displayed in Fig. \ref{fig:prior}, due to the low quality and lighting variations of the source images, we find that the detected saliency maps have noises and distortions such as the hairs, and faces on the CUFSF dataset. Considering that, we drop the saliency maps and still achieve the \textbf{best performance} on the biphasic photo-sketch synthesis results on the CUFSF dataset. 
Besides, under \textbf{normal light conditions}, the saliency detector~\cite{b21} expresses superior performance on both face photo and sketch\footnote{There is an open-sourced platform based on this saliency detector for more demos: http://profu.ai/}, thus leading to the positive impact as shown in Table \ref{tab:ablation-CUFS} and Fig. \ref{fig:CUFS}. Here, as shown in Table \ref{tab:ablation_CUFSF}, we discard the saliency maps in the following experiments after the injection of semantic layouts.

\subsubsection{Parsing Layouts Injection}
Furthermore, we adopt the semantic parsing layouts as two modulation parameters with spatial dimensions injected into the decoder of our network. The semantic layouts aim at providing a kind of spatial supervision of synthesized images. As depicted in Fig. \ref{fig:CUFS} (e) and Fig. \ref{fig:CUFSF} (d), the details of the synthesized images are more realistic and vivid in class region. The experiment results also show the effectiveness of parsing layouts injection as shown in Table \ref{tab:ablation-CUFS} and Table \ref{tab:ablation_CUFSF}.

\subsubsection{Perceptual Loss and BCE Loss}
In addition, we apply the perceptual loss to improve the high-frequency quality of synthesized images. As displayed in Table \ref{tab:ablation-CUFS} and Table \ref{tab:ablation_CUFSF}, the perceptual loss could significantly reduce the value of FID while increasing FSIM in the face sketch synthesis task. Besides, the Binary Cross-Entropy loss is implemented by BiSeNet \cite{b42} to refine the structure layouts of synthesized images. As illustrated in Fig. \ref{fig:CUFS} (g) and Fig. \ref{fig:CUFSF} (f), the synthesized human faces retain more distinct facial contours.

\subsubsection{Intra-class Semantic Graph Loss}
Moreover, we propose a novel IntrA-class semantic Graph (IASG) loss to restrain the generated images with ground truth. The IASG loss forces the synthesized images to hold more semantic intra-class knowledge. The details of the sketches and photos we generated are more similar to ground truth, such as the texture of the hair, the contour of the ears, the position of the eyebrows, and the eyes are more consistent as represented in Fig. \ref{fig:CUFS} (h) and Fig. \ref{fig:CUFSF} (g). More than that, the IASG Loss raises multiple indicators considerably such as LPIPS (alex, squeeze, vgg) as expressed in Table \ref{tab:ablation-CUFS} and Table \ref{tab:ablation_CUFSF}.

\subsubsection{Inter-class Structure Graph Loss}
Combined with the IASG loss we proposed, we bring forward a novel InteR-class structure Graph (IRSG) loss that remarkably enhances the performance of our network. As reflected in Table \ref{tab:ablation-CUFS} and Table \ref{tab:ablation_CUFSF}, the SSIM is increased in whole experiments especially on the sketch synthesis task on the CUFS dataset. The IRSG loss allows the synthesized sketches to become more structure-coordinated. It can be observed that in Fig. \ref{fig:CUFS} (i) and Fig. \ref{fig:CUFSF} (h), the sketches are more vivid and distinct.

\subsubsection{Iterative Cycle Training Loss}
Eventually, we design a novel biphasic Iterative Cycle Training (ICT) loss to boost the training of sketch and photo synthesis. There are multi-stage iterations in this training strategy. To examine the comprehensive training procedure, we express the results in Table \ref{tab:Iteration_Training}. After conducting three times iterations, the network tends to converge and reach its optimum at the third iteration. As depicted in Fig. \ref{fig:CUFS} (j) and Fig. \ref{fig:CUFSF} (i), the synthesized sketches and photos contain more personal characteristics. The details of the face images are more meticulous without less blurring and noise such as the eyelashes, hairstyles, and teeth.

\subsection{Heterogeneous Face Verification Rate}
Meanwhile, we exploit the face verification rate (FVR), deployed by the Face++ API, to report the identity preservation ability of our multi-variation models, as highlighted in Table \ref{tab:ablation-CUFS} and Table \ref{tab:ablation_CUFSF}. Our method synthesizes face photos and sketches from heterogeneous domains. Comprehensively, in the CUFS dataset, the FVR increases distinctly from the backbone to the final model by 3.332 (84.252$\rightarrow$87.542) for sketch synthesis and 8.253 (73.527$\rightarrow$81.780) for face photo synthesis. Similarly, the proposed method improves the FVR considerably despite the deformation between the paired sketch photo in the CUFSF dataset. To be more specific, the FVR increases significantly after we exploit the intra-class and inter-class representational graph loss functions, especially in the CUFSF dataset. Although there are deformations between the photos and sketches in the CUFSF dataset, the proposed graph loss functions improve the FVR by 0.511 ( 86.512$\rightarrow$87.023) for sketch synthesis and 1.026 (62.026$\rightarrow$63.052) for photo synthesis. This highlights the robustness of our proposed graph representation algorithms.

\subsection{User Study}
\begin{figure}[t]
\begin{center}
\includegraphics[width=0.95 \linewidth]{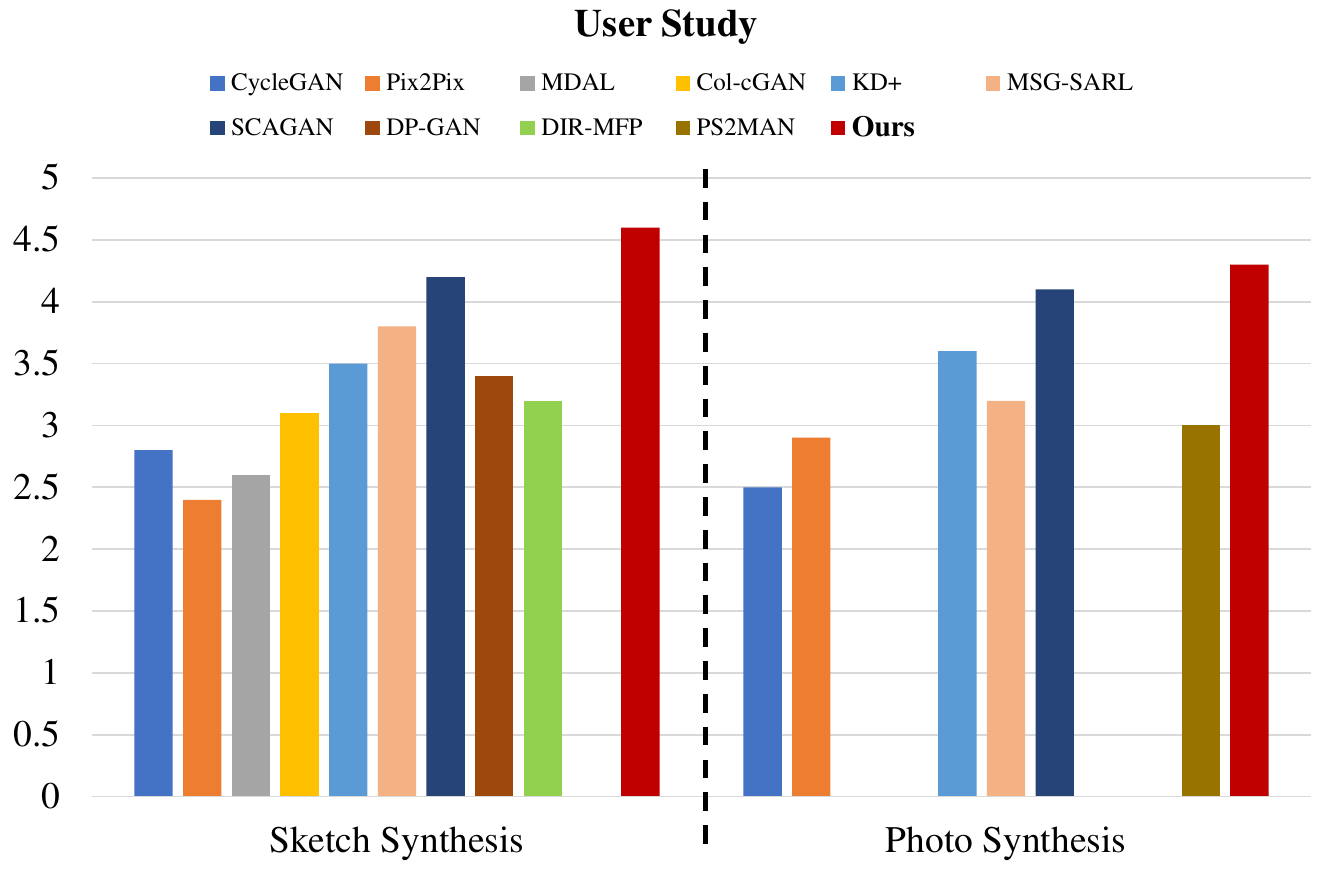}
\end{center}
\vspace{-1em}
   \caption{User study on the sketch synthesis and photo synthesis tasks. There are many approaches that can only achieve the unidirectional sketch synthesis since the photo synthesis task is more challenging.}
\label{fig:user_study}
\end{figure}

We conduct a user study to further analyze the visual quality of the generated biphasic photo-sketch from different methods. In particular, we recruit 20 volunteers and ask them to give feedback scores ranging from 0-5 (the higher, the better) of the generated faces. The evaluation perceptions consist of naturalness and authenticity. The statistical results
are reported in Fig.~\ref{fig:user_study}. Our method (\ie, the red column) achieves the best performance on both sketch and photo synthesis tasks. This strongly proves our key insight on building the two types representational graphs to model the spatial structure information of human faces.

\section{Conclusion}
In this paper, we propose a Semantic-Driven Generative Adversarial Network with Graph Representation Learning for biphasic face photo-sketch synthesis by utilizing saliency detection and face parsing layouts as prior information. In particular, parsing layouts are employed to construct two types of representational graphs that restraint the intra-class and inter-class features of the synthesized images. These graphs could enforce our network to generate the considerate face structure and details. 
In addition, based on the observation that the paired photo-
sketch shared many personal characters, we design a novel biphasic iterative cycle training strategy to refine the high-frequency quality of the synthesized images. 
Extension experiments are conducted to verify the effectiveness of proposed each module. 

Although the proposed method achieves state-of-the-art performance, there are still limitations in our work. The photos generated in the CUFSF dataset are not vivid and realistic leading to the low face verification rate of the synthesized photos. These issues are caused by the deformation between photos and sketches in the dataset. In addition, in terms of a few indicators, our method is still not optimal implying there are many improvements that can be done. In the future, we will pay more attention to the biphasic face photo-sketch synthesis when facing the large deformations between the paired human photo and sketch.


%


\ifCLASSOPTIONcaptionsoff
  \newpage
\fi

\end{document}